\documentclass[10pt,twocolumn,letterpaper]{article}

\usepackage{cvpr}
\usepackage{times}
\usepackage{epsfig}
\usepackage{graphicx}
\usepackage{amsmath}
\usepackage{amssymb}

\usepackage{times}
\usepackage{epsfig}
\usepackage{graphicx}
\usepackage{amsmath}
\usepackage{amssymb}

\usepackage{caption}

\usepackage{makecell}
\usepackage{booktabs}
\usepackage{multirow}
\usepackage{siunitx}
\usepackage{dsfont}

\usepackage[T1]{fontenc}
\usepackage{fix-cm}

\newcolumntype{?}{!{\vrule width 1pt}}
\newcolumntype{C}[1]{>{\centering}m{#1}}

\newcolumntype{X}{@{\hskip\tabcolsep\vrule width 1.5pt\hskip\tabcolsep}}

\newcommand{\myfiguretwocol}[1]{
\begin{minipage}[b]{.23\textwidth}
\includegraphics[width=1.07\linewidth]{#1}
\end{minipage}
}

\newcommand{\myfigurethreecol}[1]{
\begin{minipage}[b]{.14\textwidth}
\includegraphics[width=1.1\linewidth]{#1}
\end{minipage}
}

\usepackage[pagebackref=true,breaklinks=true,letterpaper=true,colorlinks,bookmarks=false]{hyperref}

 \cvprfinalcopy 

\def\cvprPaperID{1178} 

\ifcvprfinal\fi
\begin{document}

\title{Exploiting Egocentric Object Prior for  3D Saliency Detection}


\author{Gedas Bertasius\\
University of Pennsylvania\\
{\tt\small gberta@seas.upenn.edu}
\and
Hyun Soo Park\\
University of Pennsylvania\\
{\tt\small hypar@seas.upenn.edu}
\and
Jianbo Shi\\
University of Pennsylvania\\
{\tt\small jshi@seas.upenn.edu}
}

\maketitle


\begin{abstract}

On a minute-to-minute basis people undergo numerous fluid interactions with objects that barely register on a conscious level. Recent neuroscientific research demonstrates that humans have a fixed size prior for salient objects. This suggests that a salient object in 3D undergoes a consistent transformation such that people's visual system perceives it with an approximately fixed size. This finding indicates that there exists a consistent egocentric object prior that can be characterized by shape, size, depth, and location in the first person view.

In this paper, we develop an EgoObject Representation, which encodes these characteristics by incorporating shape, location, size and depth features from an egocentric RGBD image. We empirically show that this representation can accurately characterize the egocentric object prior by testing it on an egocentric RGBD dataset for three tasks: the 3D saliency detection, future saliency prediction, and interaction classification. This representation is evaluated on our new Egocentric RGBD Saliency dataset that includes various activities such as cooking, dining, and shopping. By using our EgoObject representation, we outperform previously proposed models for saliency detection (relative $30\%$ improvement for 3D saliency detection task) on our dataset. Additionally, we demonstrate that this representation allows us to predict future salient objects based on the gaze cue and classify people's interactions with objects. 




\end{abstract}


%
%
%
%
%
%

\section{Introduction}

\begin{figure}
\begin{center}
   \includegraphics[width=1.05\linewidth]{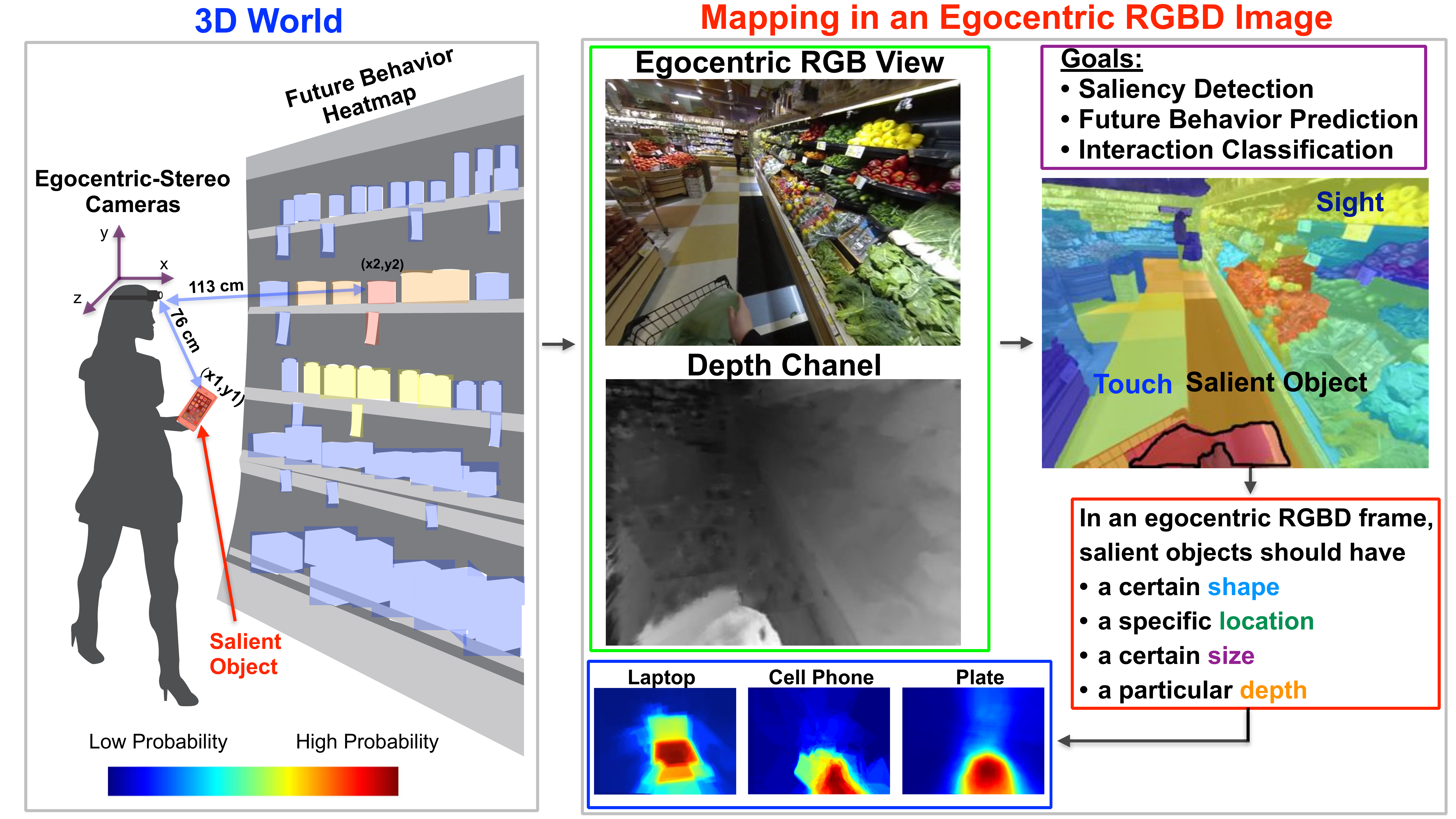}
\end{center}
\caption{An illustration of our approach (best viewed in color). Our goal is to use egocentric-stereo cameras to characterize an egocentric object prior in a first-person view RGBD frame and use it for 3D saliency detection. Based on recent psychology findings indicating that humans have a fixed size prior for salient objects, we conjecture that an object from 3D world should map to an egocentric RGBD image with some predictable shape, size, location and depth pattern. Using this intuition, we propose an EgoObject representation that encodes these characteristics in an egocentric RGBD frame. We then show its effectiveness on our collected egocentric RGBD Saliency dataset for the tasks of 3D saliency detection, future saliency prediction and interaction classification.}
\label{fig:main}
\end{figure}

On a daily basis, people undergo numerous interactions with objects that barely register on a conscious level. For instance, imagine a person shopping at a grocery store as shown in Figure~\ref{fig:main}. Suppose she picks up a can of juice to load it in her shopping cart. The distance of the can is maintained fixed due to the constant length of her arm. When she checks the expiration date on the can, the distance and orientation towards the can is adjusted with respect to her eyes so that she can read the label easily. In the next aisle, she may look at a LCD screen at a certain distance to check the discount list in the store. Thus, this example shows that spatial arrangement between objects and humans is subconsciously established in 3D. In other words, even though people do not consciously plan to maintain a particular distance and orientation when interacting with various objects, these interactions usually have some consistent pattern. This suggests the existence of an egocentric object prior in the person's field of view, which implies that a 3D salient object should appear at a predictable location, orientation, depth, size and shape when mapped to an egocentric RGBD image. 

Our main conjecture stems from the recent work on human visual perception~\cite{stroop}, which shows that {\em humans possess a fixed size prior for salient objects}. This finding suggests that a salient object in 3D undergoes a transformation such that people's visual system perceives it with an approximately fixed size. Even though, each person's interactions with the objects are biased by a variety of factors such as hand dominance or visual acuity, common trends for interacting with objects certainly exist. In this work, we investigate whether one can discover such consistent patterns by exploiting egocentric object prior from the first-person view in RGBD frames.


Our problem can be viewed as an inverse object affordance task~\cite{hall_humans,koppula2015_anticipatingactivities,fathi}. While the goal of a traditional object affordance task is to predict human behavior based on the object locations, we are interested in predicting potential salient object locations based on the human behavior captured by an egocentric RGBD camera. The core challenge here is designing a representation that would encode generic characteristics of visual saliency without explicitly relying on object class templates~\cite{PirsiavashR_CVPR_2012_1} or hand skin detection~\cite{fathi}. Specifically, we want to design a representation that captures how a salient object in the 3D world, maps to an egocentric RGBD image.  Assuming the existence of an egocentric object prior in the first-person view, we hypothesize that a 3D salient object would map to an egocentric RGBD image with a predictable shape, location, size and depth pattern. Thus, we propose an EgoObject representation that represents each region of interest in an egocentric RGBD video frame by its  {\em shape}, {\em location}, {\em size}, and {\em depth}. Note that using egocentric camera in this context is important because it approximates the person's gaze direction and allows us to see objects from a first-person view, which is an important cue for saliency detection. Additionally,  depth information is also beneficial because it provides an accurate measure of object's distance to a person. We often interact with objects using our hands (which have a fixed length), which suggests that depth defines an important cue for saliency detection as well.  Thus assuming the existence of an egocentric object prior, our EgoObject representation should allow us to accurately predict pixelwise saliency maps in egocentric RGBD frames. 



To achieve our goals, we create a new egocentric RGBD Saliency dataset. Our dataset captures people's interactions with objects during various activities such as shopping, cooking, dining. Additionally, due to the use of egocentric-stereo cameras, we can accurately capture depth information of each scene. Finally we note that  our dataset is annotated for the following three tasks: saliency detection, future saliency prediction, and interaction classification. We show that we can successfully apply our proposed egocentric representation on this dataset and achieve solid results for these three tasks. These results demonstrate that by using our EgoObject representation, we can accurately characterize an egocentric object prior in the first-person view RGBD images, which implies that salient objects from the 3D world map to an egocentric RGBD image with predictable characteristics of shape, location, size and depth. We demonstrate that we can learn this egocentric object prior from our dataset and then exploit it for 3D saliency detection in egocentric RGBD images.




\captionsetup{labelformat=default}
\captionsetup[figure]{skip=10pt}

\begin{figure}
\begin{center}
   \includegraphics[width=1\linewidth]{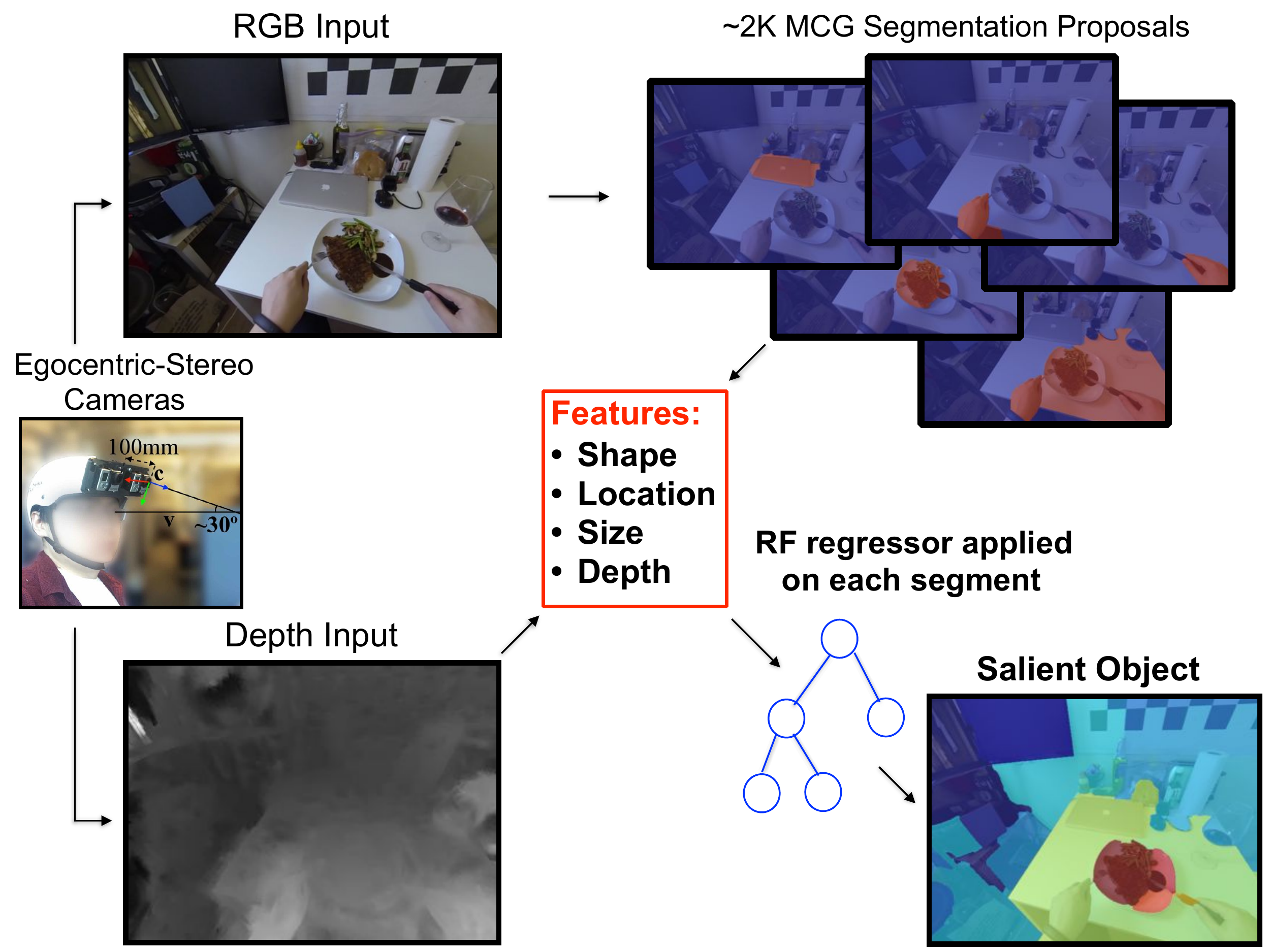}
\end{center}
\caption{An illustration of our technical approach (best viewed in color). Using egocentric-stereo cameras we record an egocentric RGBD view of a scene. We then utilize MCG~\cite{APBMM2014} method to generate $\approx 2K$ region proposals. For each of the regions $R_i$ we then generate a feature vector $x(R_i)$ that captures shape, location, size and depth cues and use these features to predict the 3D saliency of region $R_i$.}
\label{fig:method}
\end{figure}




\section{Related Work}

\textbf{Saliency Detection in Images.} In the past, there has been much research on the task of saliency detection in 2D images. Some of the earlier work employs bottom-up cues, such as color, brightness, and contrast to predict saliency in images~\cite{Harel07graph-basedvisual,yang2013saliency,Itti:1998:MSV:297843.297870,Achanta_frequency-tunedsalient}. Additionally, several methods demonstrate the importance of shape cues for saliency detection task~\cite{Jiang11automaticsalient,DBLP:journals/corr/LiHKRY14}. Finally, some of the more recent work employ object-proposal methods to aid this task~\cite{objectness,cpmc-release1,APBMM2014}.


Unlike the above listed methods that try to predict saliency based on contrast, brightness or color cues, we are more interested in expressing an egocentric object prior based on shape, location, size and depth cues in an egocentric RGBD image. Our goal is then to use such prior for 3D saliency detection in the egocentric RGBD images.


\textbf{Egocentric Visual Data Analysis.} In the recent work, several methods employed egocentric (first-person view) cameras for the tasks such as video summarization~\cite{DBLP:journals/ijcv/LeeG15,Lu:2013:SSE:2514950.2516026}, video stabilization~\cite{Kopf:2014:FHV:2601097.2601195}, object recognition~\cite{conf/cvpr/RenG10,DBLP:journals/corr/BolanosR15}, and action and activity recognition~\cite{PirsiavashR_CVPR_2012_1,Fathi:2011:UEA:2355573.2356302,SpriggsDH09,Li_2015_CVPR}.

In comparison to the prior egocentric approaches we propose a novel problem, which can be formulated as an inverse object affordance problem: our goal is to detect 3D saliency in egocentric RGBD images based on human behavior that is captured by egocentric-stereo cameras. Additionally, unlike prior approaches, we use \textbf{egocentric-stereo} cameras to capture egocentric RGBD data. In the context of saliency detection, the depth information is important because it allows us to accurately capture object's distance to a person. Since people often use hands (which have fixed length) to interact with objects, depth information defines an important cue for saliency detection in egocentric RGBD environment. 

Unlike other methods, which rely on object detectors~\cite{PirsiavashR_CVPR_2012_1}, or hand and skin segmentation~\cite{fathi,DBLP:journals/ijcv/LeeG15}, we propose EgoObject representation that is based solely on shape, location, size and depth cues in an egocentric RGBD images. We demonstrate that we can use our representation successfully to predict 3D saliency in egocentric RGBD images.

\section{EgoObject Representation}
\label{feats}



Based on our earlier hypothesis, we conjecture that objects from the 3D world map to an egocentric RGBD image with some predictable {\em shape}, {\em location}, {\em size} and {\em depth}. We encode such characteristics in a region of interest $\mathcal{R}$ using an EgoObject map, $f(\mathcal{R}) = \left[\begin{array}{cc}\phi(\mathcal{R})^\mathsf{T}&\xi(\mathcal{R})^\mathsf{T}\end{array}\right] \in \mathds{R}^{N_s\times N_l \times N_b \times N_d \times N_c}$ where $N_s$, $N_l$, $N_b$, $N_d$, and $N_c$ are the number of the feature dimension for shape $\phi_s$, location $\phi_l$, size $\phi_b$, depth $\phi_d$, and context $\xi$, respectively.




\captionsetup{labelformat=empty}
\captionsetup[figure]{skip=10pt}

\begin{figure}
\centering

\myfigurethreecol{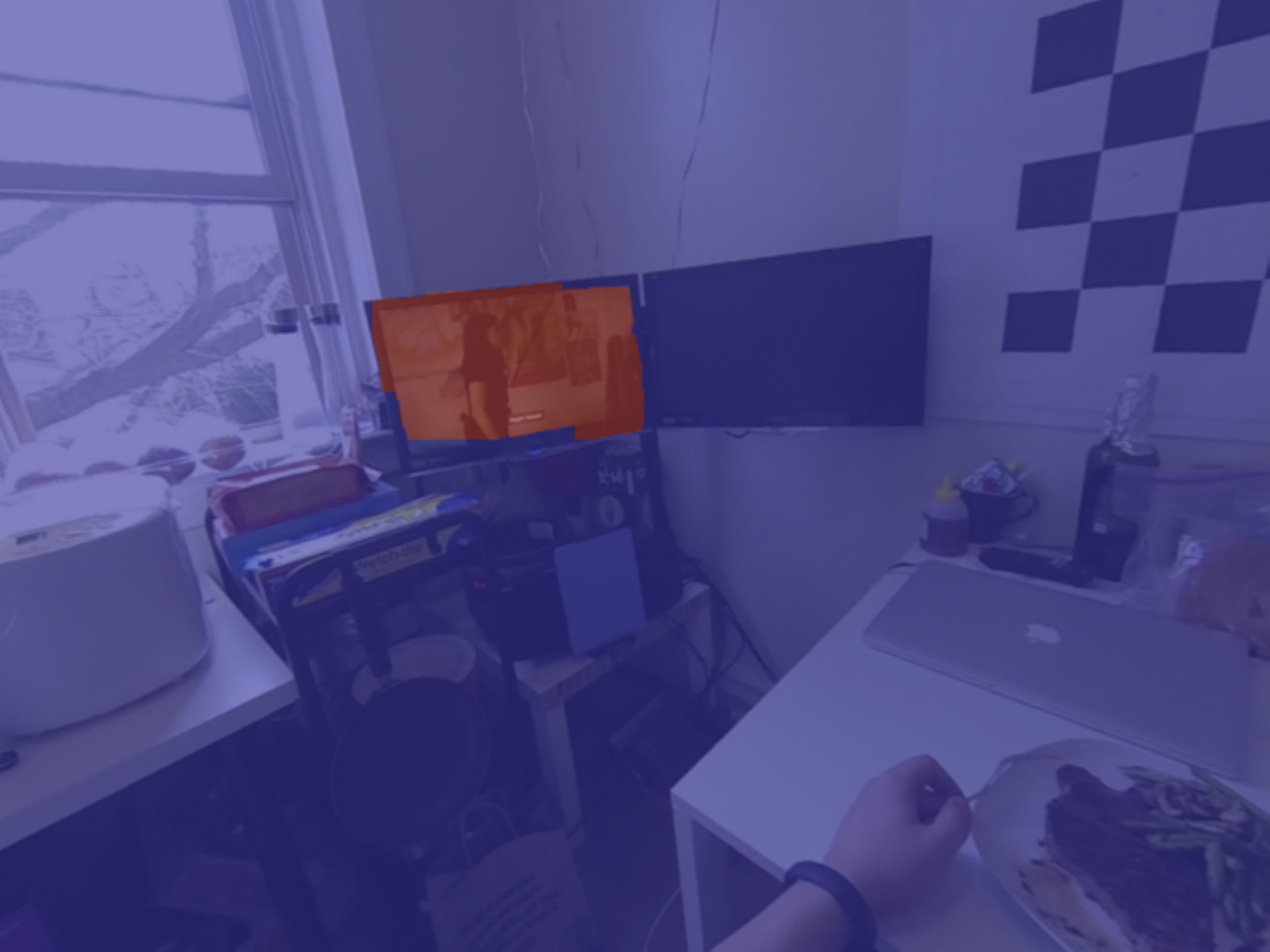}
\myfigurethreecol{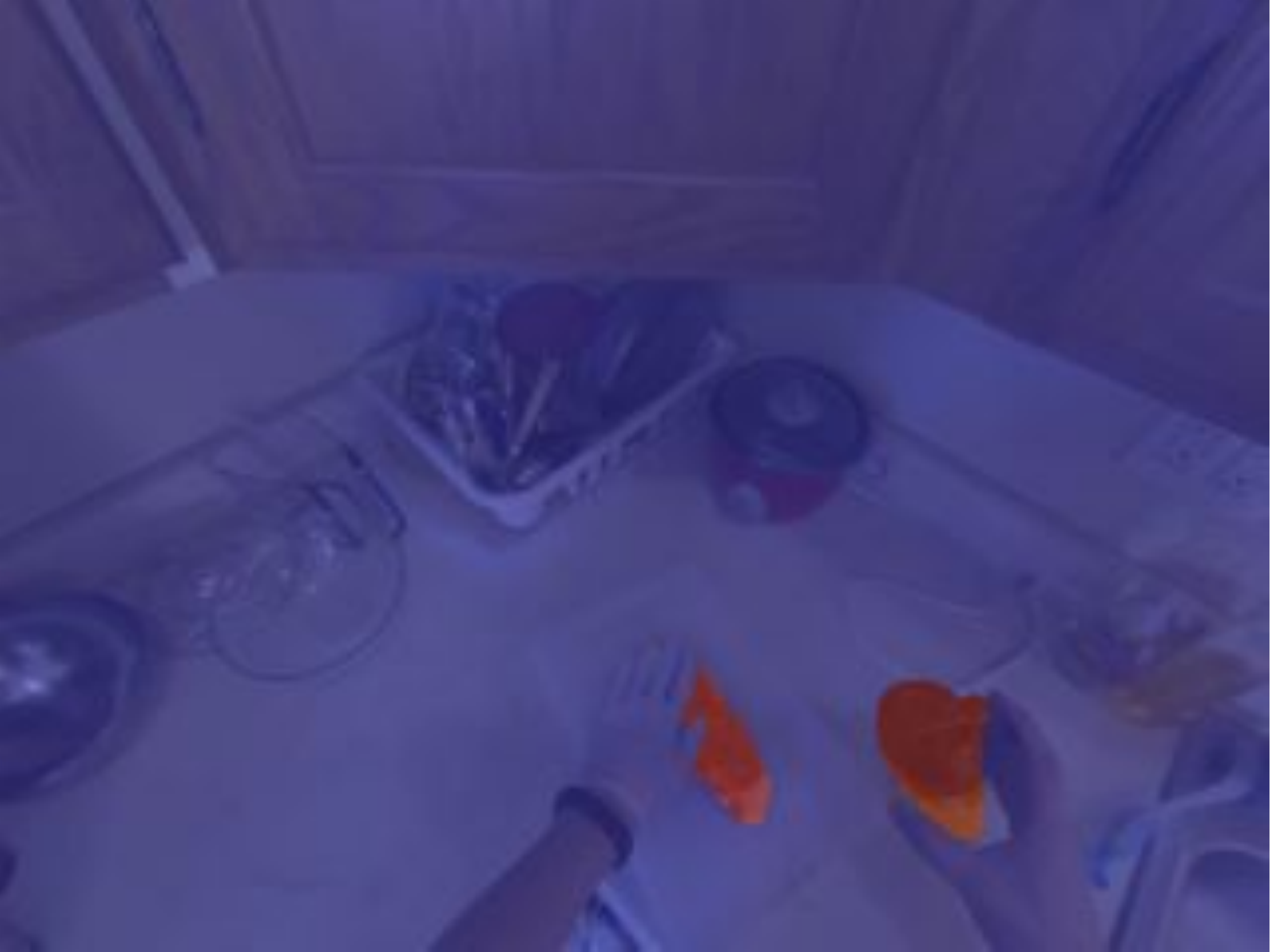}
\myfigurethreecol{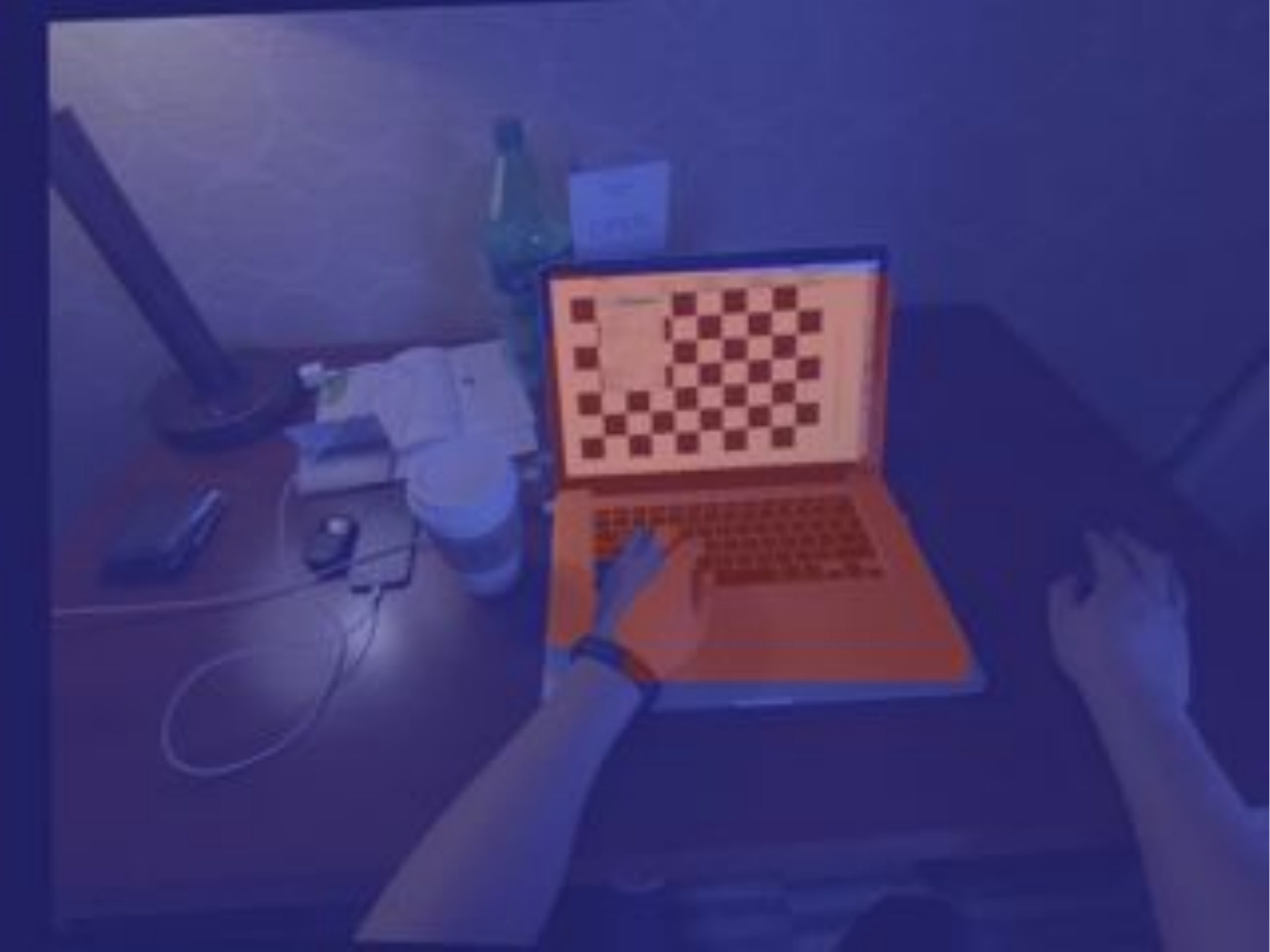}

\myfigurethreecol{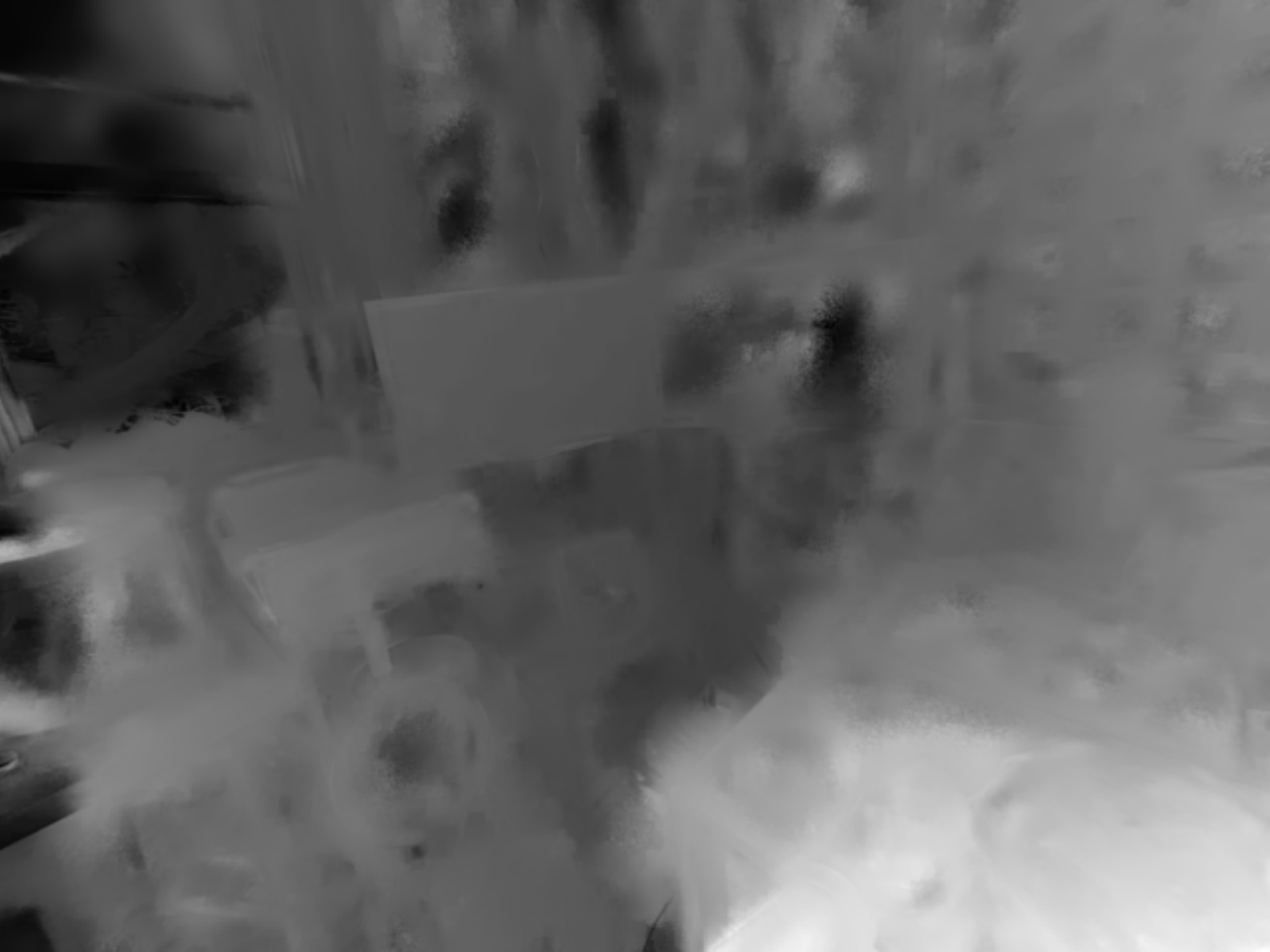}
\myfigurethreecol{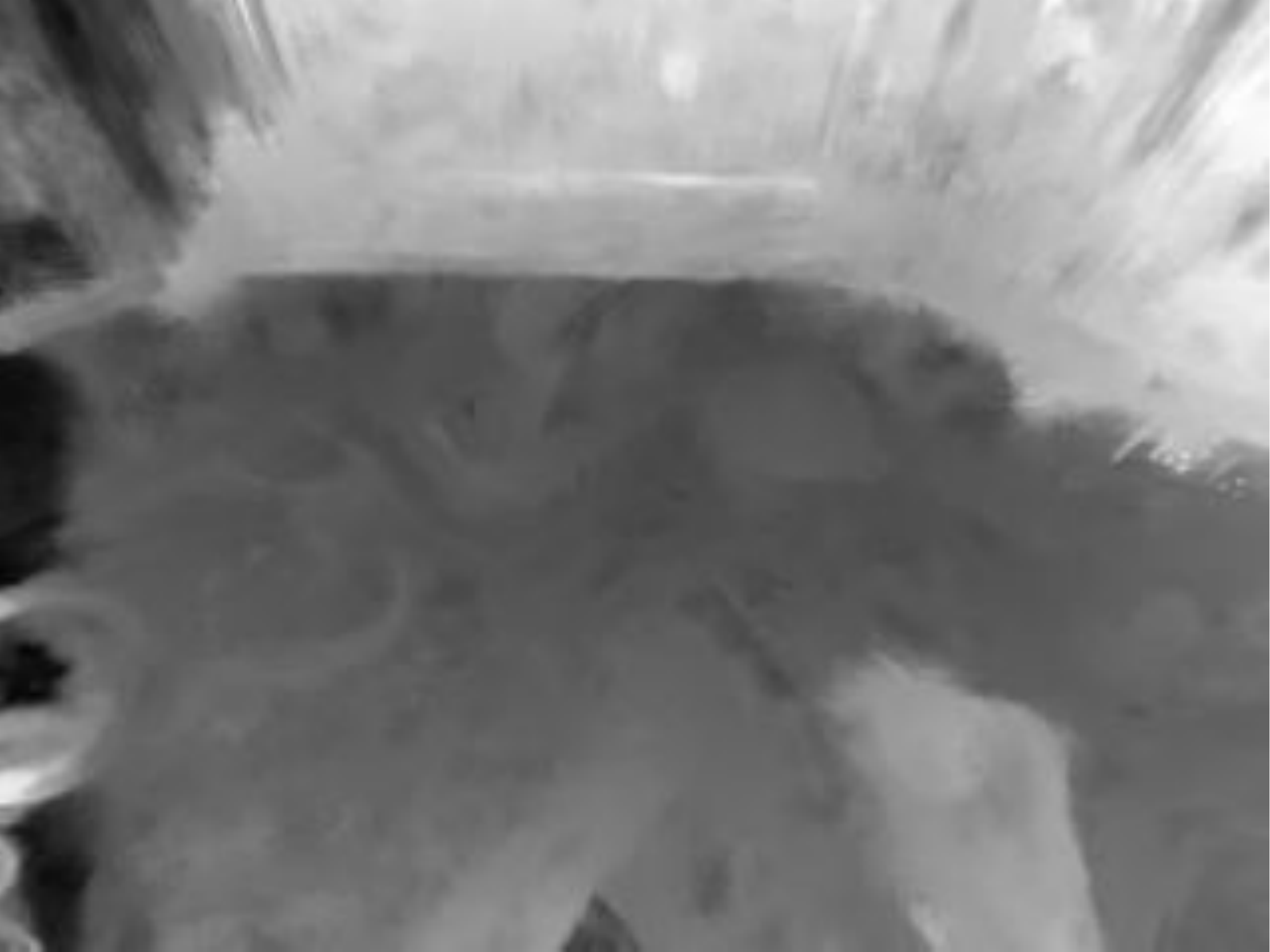}
\myfigurethreecol{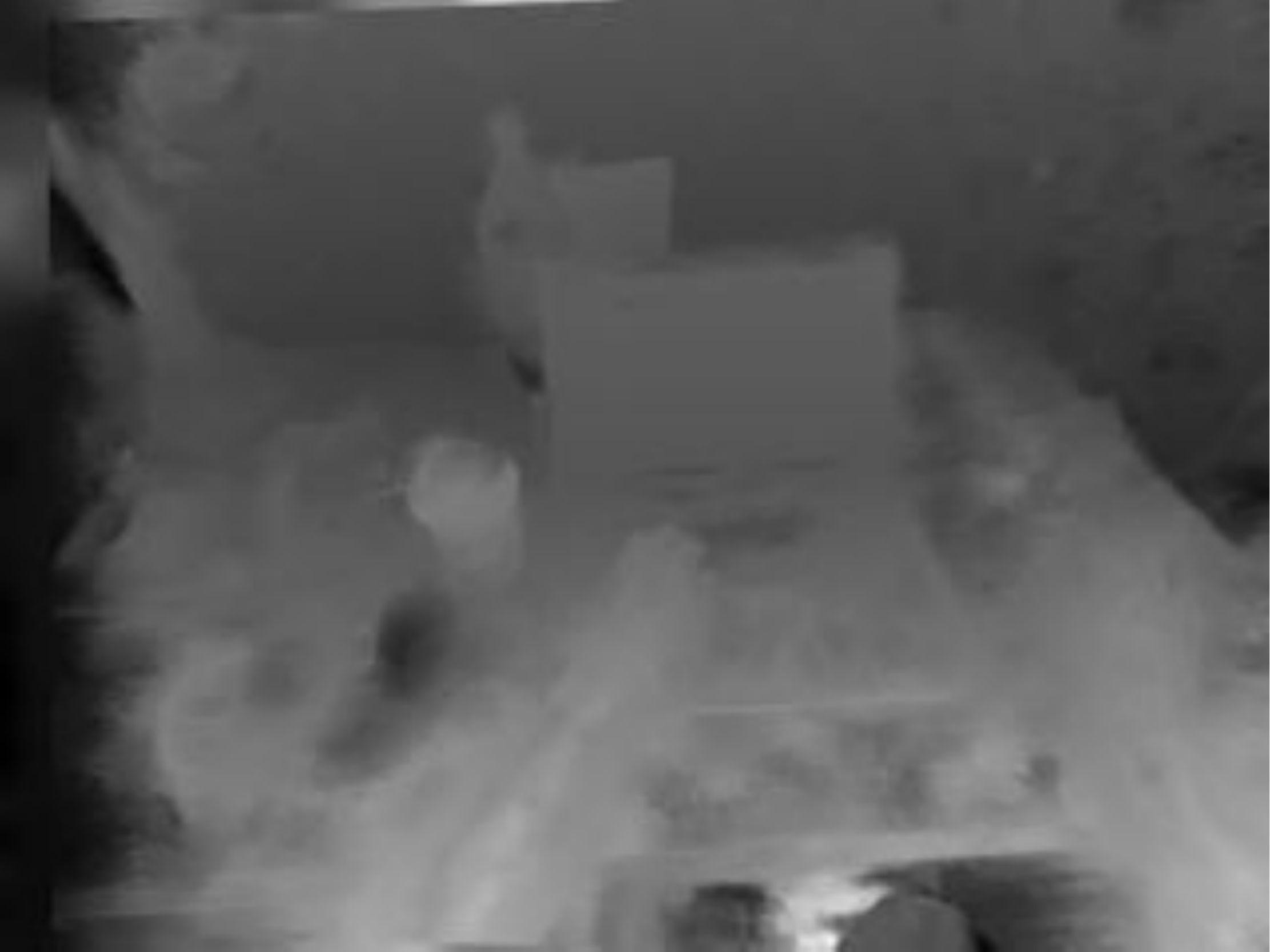}

\captionsetup{labelformat=default}
    \caption{A figure illustrating a few sample images from our Egocentric RGBD Saliency dataset (best viewed in color). The top row displays RGB channels overlaid with the salient object annotations, whereas the bottom row shows corresponding depth maps.}
    \label{fig:data_po}
\end{figure}


A shape feature, $\phi_s(\mathcal{R}) = \left[\begin{array}{ccc}\phi_{s_1}^\mathsf{T}&\phi_{s_2}^\mathsf{T}&\phi_{s_3}^\mathsf{T}\end{array}\right]^\mathsf{T}$ captures a geometric properties such as area, perimeter, edges, and orientation of $\mathcal{R}$. 
\begin{itemize}
\item $\phi_{s_1}(\mathcal{R}) \in \mathds{R}^4$: perimeter divided by the squared root of the area, the area of a region divided by the area of the bounding box, major and minor axes lengths. \vspace{-2mm} 
\item $\phi_{s_2}(\mathcal{R}) \in \mathds{R}^4$: we employ boundary cues~\cite{Dollar2015PAMI}, which include, sum and average contour strength of boundaries in region $\mathcal{R}$ and also minimum and maximum ultrametric-contour values that lead to appearance and disappearance of the smaller regions inside $\mathcal{R}$~\cite{Arbelaez:2011:CDH:1963053.1963088}. \vspace{-2mm} 
\item $\phi_{s_3}(\mathcal{R}) \in \mathds{R}^3$: eccentricity and orientation of $\mathcal{R}$ and also the diameter of a circle with the same area as region $\mathcal{R}$.
\end{itemize}

A location feature $\phi_l(\mathcal{R}) = \left[\begin{array}{cc}\phi_{l_1}^\mathsf{T}&\phi_{l_2}^\mathsf{T}\end{array}\right]^\mathsf{T}$ encode spatial prior of objects imaged in an egocentric view:
\begin{itemize}
\item $\phi_{l_1}(\mathcal{R}) \in \mathds{R}^6$: normalized bounding box coordinates and the centroid of a region $\mathcal{R}$. \vspace{-2mm}
\item $\phi_{l_2}(\mathcal{R}) \in \mathds{R}^{10}$: we also compute horizontal and vertical distances from the centroid of $R_i$ to the center of an image, and also to the mid-points of each border in the image.
\end{itemize}

A size feature $\phi_b(\mathcal{R}) = \left[\begin{array}{cc}\phi_{b_1}^\mathsf{T}&\phi_{b_2}^\mathsf{T}\end{array}\right]^\mathsf{T}$ encodes the size of the bounding box and area of the region.
\begin{itemize}
\item $\phi_{b_1}(\mathcal{R}) \in \mathds{R}^2$: area and perimeter of region $\mathcal{R}$. \vspace{-2mm}
\item $\phi_{b_2}(\mathcal{R}) \in \mathds{R}^2$: area and aspect ratio of the bounding box corresponding to the region $\mathcal{R}$.
\end{itemize}

$\phi_d(\mathcal{R}) = \left[\begin{array}{ccccc}\phi_{d_1}^\mathsf{T}&\phi_{d_2}^\mathsf{T}&\phi_{d_3}^\mathsf{T}&\phi_{d_4}^\mathsf{T}&\phi_{d_5}^\mathsf{T}\end{array}\right]^\mathsf{T}$ encodes a spatial distribution depth within $\mathcal{R}$.
\begin{itemize}
\item $\phi_{d_1}(\mathcal{R}) \in \mathds{R}^4$: minimum, average, maximum, depth and also standard deviation of depth in a region $\mathcal{R}$. \vspace{-2mm}
\item $\phi_{d_2}(\mathcal{R}) \in \mathds{R}^9$: $3 \times 3$  spatial depth histograms over the region $\mathcal{R}$.\vspace{-2mm}
\item $\phi_{d_3}(\mathcal{R}) \in \mathds{R}^{12}$: $4 \times 3$  depth histograms over the region $\mathcal{R}$ aligned to its major axis.\vspace{-2mm}
\item $\phi_{d_4}(\mathcal{R}) \in \mathds{R}^9$: $3 \times 3$  spatial {\em normalized} depth histograms over the region $\mathcal{R}$.\vspace{-2mm}
\item $\phi_{d_5}(\mathcal{R}) \in \mathds{R}^{12}$: $4 \times 3$  {\em normalized} depth histograms over the region $\mathcal{R}$ aligned to its major axis.
\end{itemize}



In addition, we include a context feature $\xi$ that encodes a spatial relationship between near regions in the egocentric image. Given two regions, $\mathcal{D}_{l:m}(\mathcal{R}_i,\mathcal{R}_j)$ computes a distance between two features, i.e., 
\begin{align}
\mathcal{D}_{l:m} (\phi(\mathcal{R}_i),\phi(\mathcal{R}_j)) = \left[\begin{array}{c}|\phi_l(\mathcal{R}_i)-\phi_l(\mathcal{R}_j)|\\ \vdots \\|\phi_m(\mathcal{R}_i)-\phi_m(\mathcal{R}_j)|\end{array}\right]^\mathsf{T}.\nonumber
\end{align} 
Given a target region, $\mathcal{R}$, the context feature $\xi(\mathcal{R}) = \left[\begin{array}{cccc}\xi_{1}^\mathsf{T}&\xi_{2}^\mathsf{T}&\xi_{3}^\mathsf{T}&\xi_{4}^\mathsf{T}\end{array}\right]^\mathsf{T}$ computes the relationship with $n$ neighboring regions, $\{\mathcal{R}_i\}_{i=1}^n$:
\begin{itemize}
\item $\xi_{1}(\mathcal{R}) \in \mathds{R}^{78}$: $\mathcal{D}_{1:78}(\phi(\mathcal{R}), \phi_{\rm min})$ \vspace{-2mm}
\item $\xi_{2}(\mathcal{R}) \in \mathds{R}^{78}$: $\mathcal{D}_{1:78}(\phi(\mathcal{R}), \phi_{\rm mean})$ \vspace{-2mm}
\item $\xi_{3}(\mathcal{R}) \in \mathds{R}^{78}$: $\mathcal{D}_{1:78}(\phi(\mathcal{R}), \phi_{\rm max})$ \vspace{-2mm}
\item $\xi_{4}(\mathcal{R}) \in \mathds{R}^{78\times k}$: $\{\mathcal{D}_{1:78}(\phi(\mathcal{R}), \phi(\mathcal{R}_{\rm knn}))\}_k$
\end{itemize}
where $\phi_{\rm min}$ and $\phi_{\rm max}$ are the feature vector constructued by the min-pooling and max-pooling of neighboring regions for each dimension. $\phi_{\rm mean}$ takes average of neighboring features and $\phi(\mathcal{R}_{\rm knn})$ is the feature of the top $k^{\rm th}$ nearest neighbor.


\textbf{Summary.}  For every region of interest $\mathcal{R}$ in an egocentric RGBD frame, we produce a $1089$ dimensional feature vector denoted by $f(\mathcal{R})$. We note that some of these features have been successfully used previously in tasks other than 3D saliency detection~\cite{APBMM2014,secrets2014li}.  Additionally, observe that we do not use any object-level feature or hand or skin detectors as is done~\cite{PirsiavashR_CVPR_2012_1, fathi,Lu:2013:SSE:2514950.2516026,DBLP:journals/ijcv/LeeG15}. This is because, in this work, we are primarily interested in studying the idea that salient objects from the 3D world are mapped to an egocentric RGBD frame with a consistent shape, location, size and depth patterns. We encode these cues with our EgoObject representation and show its effectiveness on egocentric RGBD data in the later sections of the paper.



\section{Prediction}
\label{tech_approach}

Given an RGBD frame as an input to our problem, we first feed RGB channels to an MCG~\cite{APBMM2014} method, which generates $\approx 2K$ proposal regions. Then, for each of these regions $\mathcal{R}$, we generate our proposed features $f(\mathcal{R})$ and use it as an input to the random forest classifier (RF). Using a RF, we aim to learn the function that takes the feature vector $f(\mathcal{R})$ corresponding to a particular region $\mathcal{R}$ as an input, and produces an output for one our proposed tasks for region $\mathcal{R}$ (i.e. saliency value or interaction classification). We can formally write this function as $G:\mathds{R}^{1089}\rightarrow\mathds{R}$.

We apply the following pipeline for the following three tasks: 3D saliency detection, future saliency prediction, and interaction classification. However, for each of these tasks we define a different output objective $G(f(\mathcal{R}))$ and train RF classifier according to that objective separately for each task. Below we describe this procedure for each task in more detail.

\textbf{3D Saliency Detection.} We train a random forest {\em regressor} to predict region's $\mathcal{R}$ Intersection over Union (IOU) with a ground truth salient object. To train the RF regressor we sample $\approx 70K$ regions from our dataset, and extract our features from each of these regions. We then assign a corresponding ground truth IOU value to each of them and train a RF regressor using $50$ trees. Our RF learns the mapping $G:\mathds{R}^{1089}\rightarrow[0,1]$ where $G(f(\mathcal{R}))$ denotes the ground truth IOU value between the $\mathcal{R}$ and the ground truth salient object. To deal with the imbalance issue, we sample an equal number of examples corresponding to the IOU values of $[0,0.25],[0.25,0.5],[0.5,0.75]$, and $[0.75,1]$. 


At testing time, we use MCG~\cite{APBMM2014} to generate $\approx 2K$ regions of interest. We then apply our trained RF for every region $\mathcal{R}$ and predict $\hat{y_i}$, which denotes the saliency of a region $\mathcal{R}$.  We note that MCG produces the set of regions that overlap with each other. Thus, for the pixels belonging to multiple overlapping regions $\mathcal{R}_i \hdots \mathcal{R}_k$, we assign a saliency value that corresponds to the maximum predicted value across the overlapping regions (i.e. $\max{\{\hat{y_i} \hdots \hat{y_k}\}}$). We illustrate the basic pipeline of our approach in Fig.~\ref{fig:method}.

\captionsetup{labelformat=empty}
\captionsetup[figure]{skip=10pt}

\begin{figure}
\centering

 \myfiguretwocol{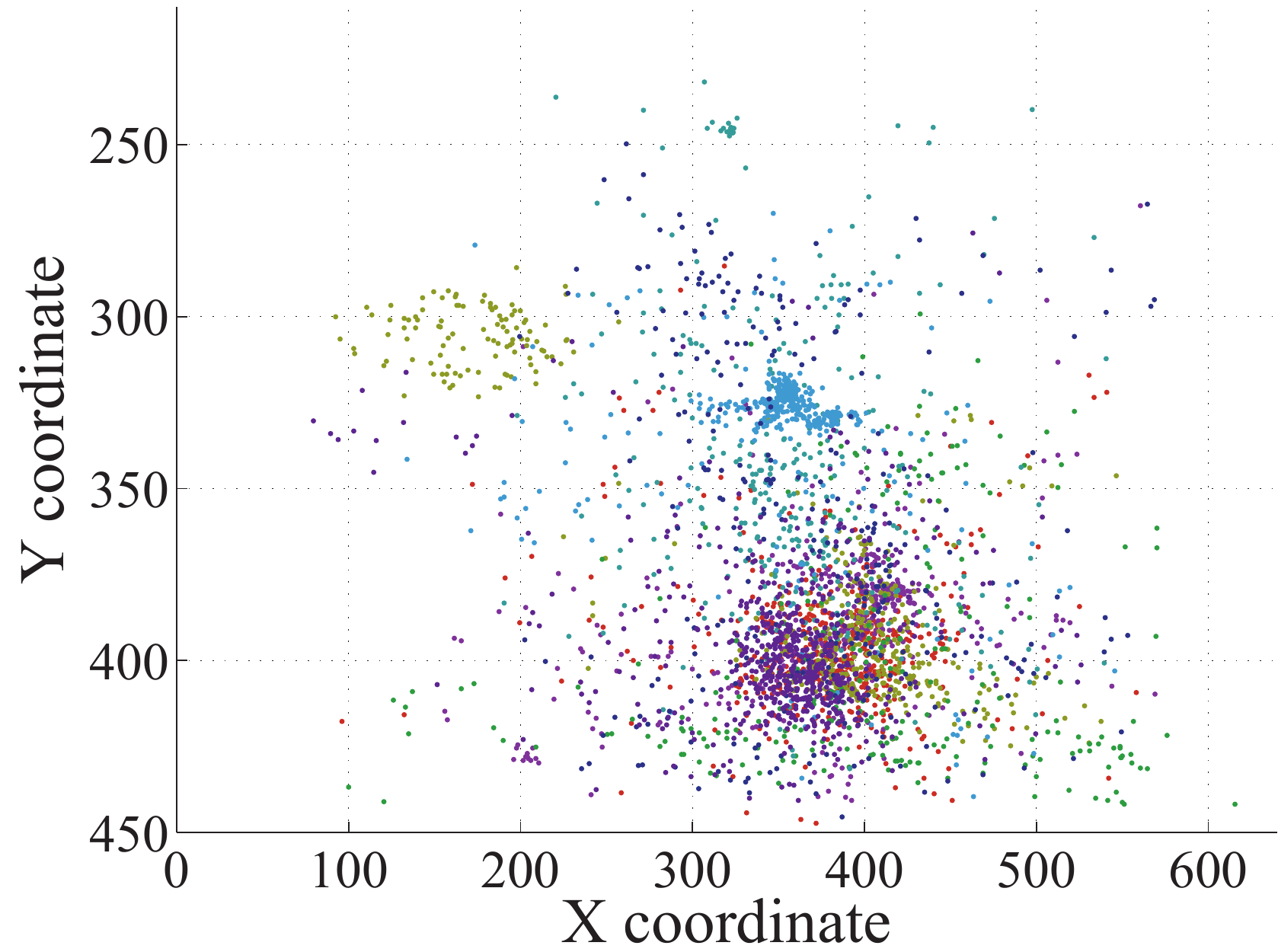}
\myfiguretwocol{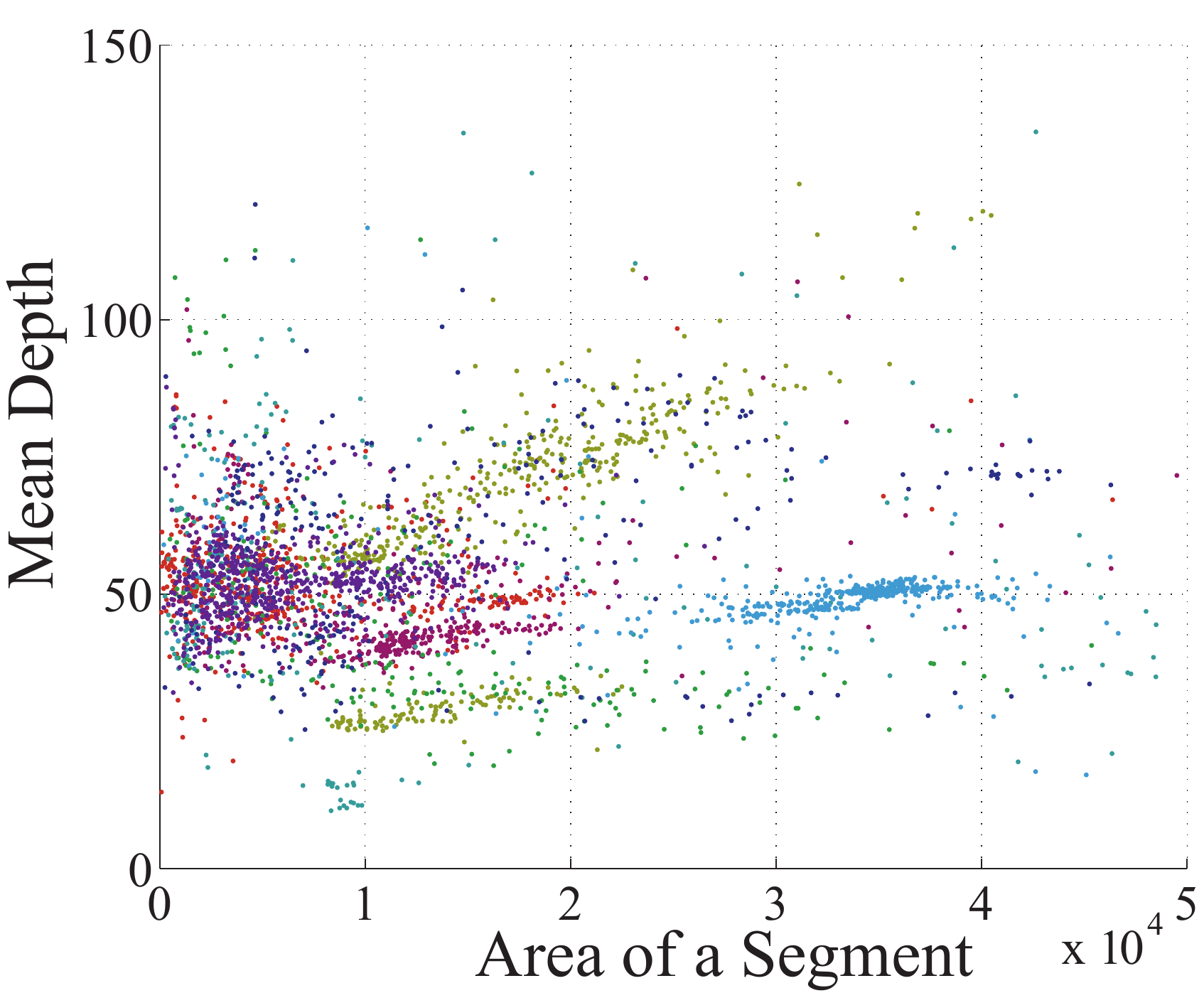}

  \captionsetup{labelformat=default}
    \caption{A figure illustrating some of the key statistics of our Egocentric RGBD Saliency dataset (best viewed in color).  Each video sequence from our dataset is marked by a different color in this plot. The figure on the left illustrates the ground truth salient object locations in an egocentric frame for all sequences. The figure on the right plots the mean depth of a salient object with respect to its area for all sequences. These figures suggest that different sequences in our dataset capture a diverse set of aspects related to people's interactions with objects.}
    \label{fig:data_stats}
\end{figure}

\textbf{Future Saliency Prediction.} For future saliency prediction, given a video frame, we want to predict, which object will be salient (i.e. used by a person) after $K$ seconds. We hypothesize that the gaze direction is one of the most informative cues that are indicative of person's future behavior. However, gaze signal may be noisy if we consider only a single frame in the video. For instance, this may happen due to the person's attention being focused somewhere else for a split second or due to the shift in the camera.

To make our approach more robust to the fluctuations of person's gaze, we incorporate simple temporal features into our system. Our goal is to use these temporal cues to normalize the gaze direction captured by an egocentric camera and make it more robust to the small camera shifts.

Thus, given a frame $F_t$ which encodes time $t$, we also consider frames $F_{t-5 \hdots t-1}$. We pair up each of these frames $F_{t-k}$ with $F_t$ and compute their respective homography matrix $H^t_{t-k}$. We then use each $H^t_{t-k}$ to recompute the image center $(C^k_x,C^k_y)$ in the current frame $F_t$. For every region $R_i$ we then recompute its distance $d^k_i$ to the new center  $(C^k_x,C^k_y)$ for all $k \in{[1,5]}$ and concatenate these new distances to the original features $f(\mathcal{R})$. Such gaze normalization scheme ensures robustness to our system in the case of gaze fluctuations.

\captionsetup{labelformat=empty}

\begin{figure}
\centering

\myfigurethreecol{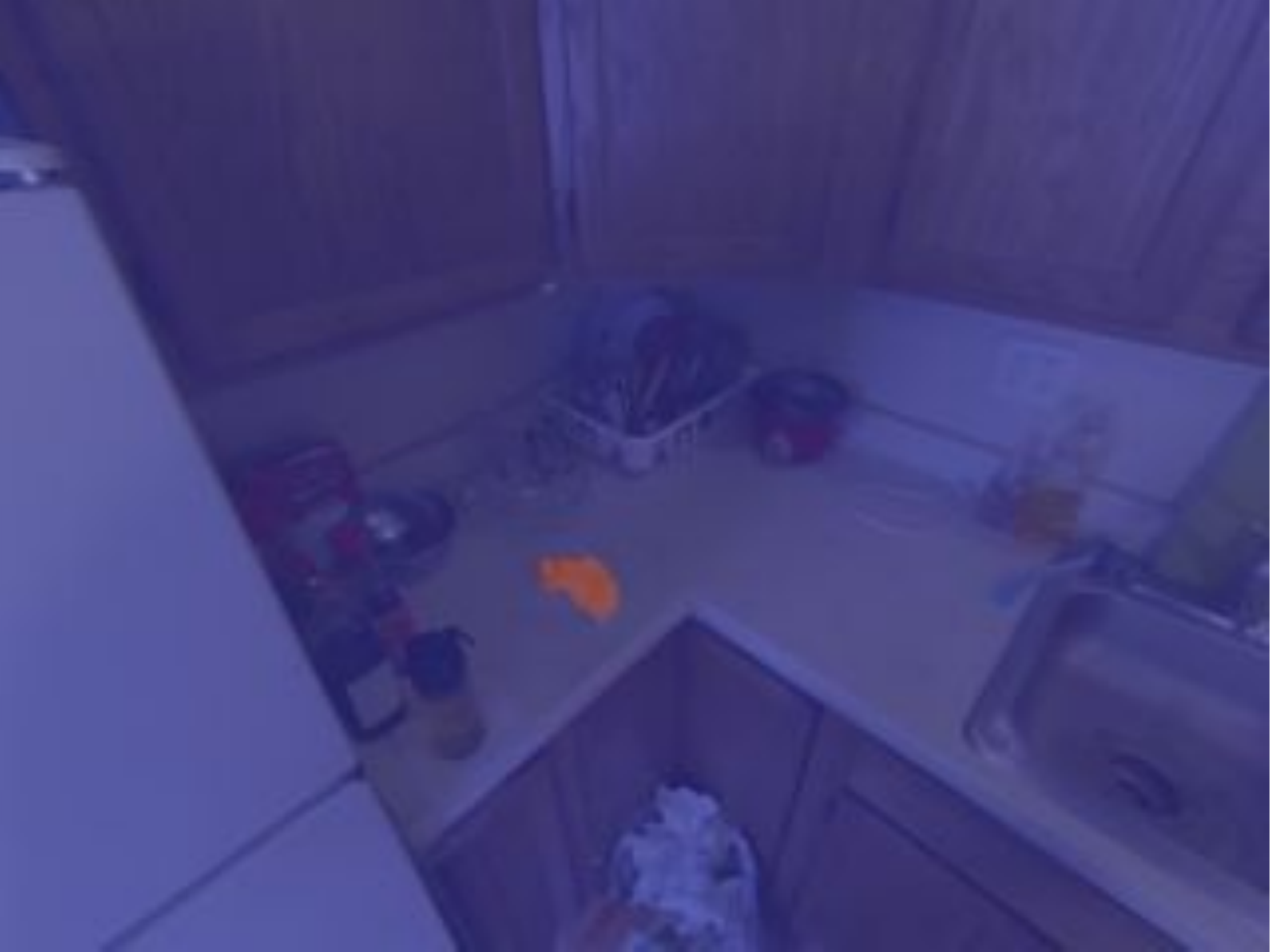}
\myfigurethreecol{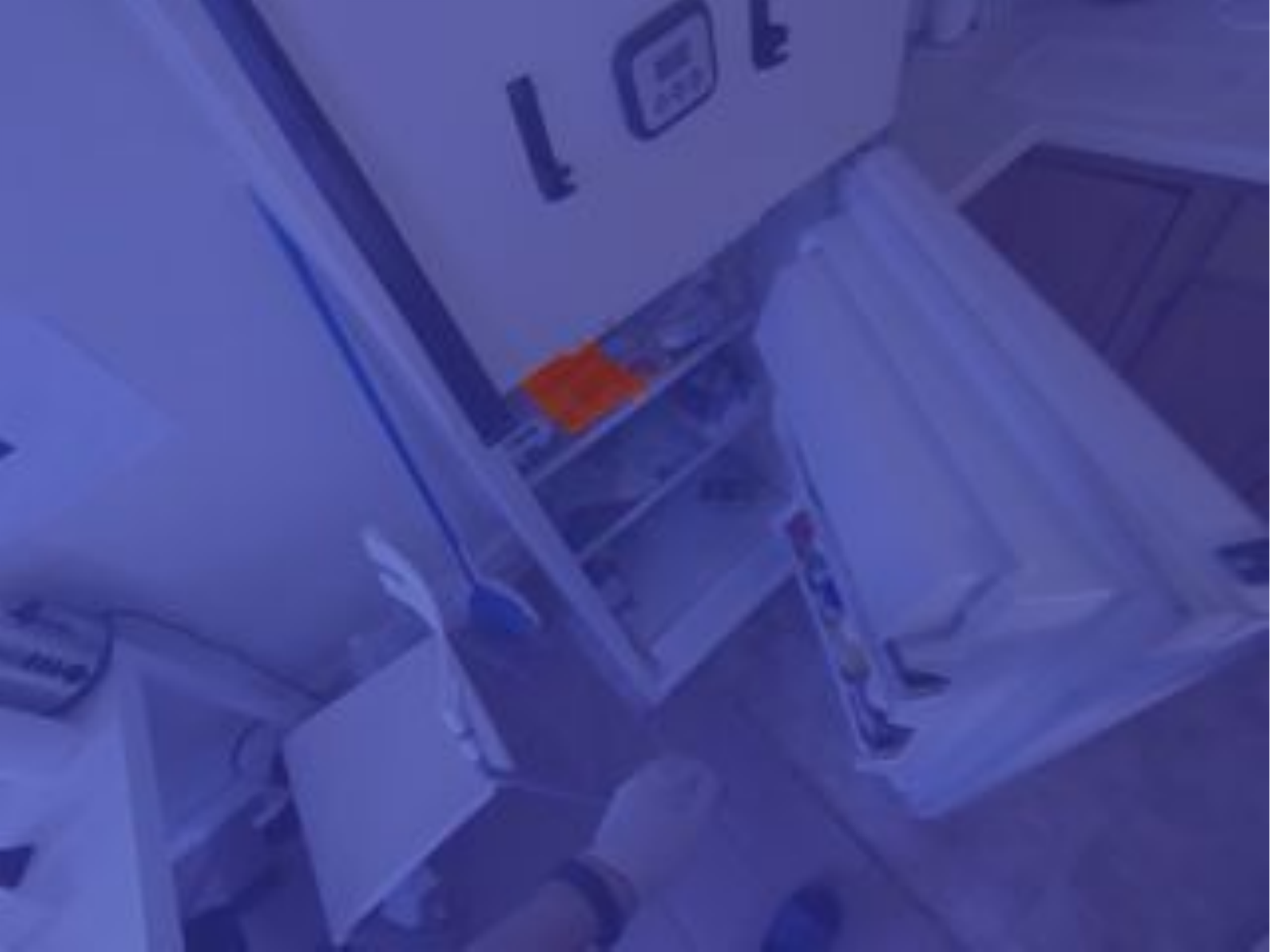}
\myfigurethreecol{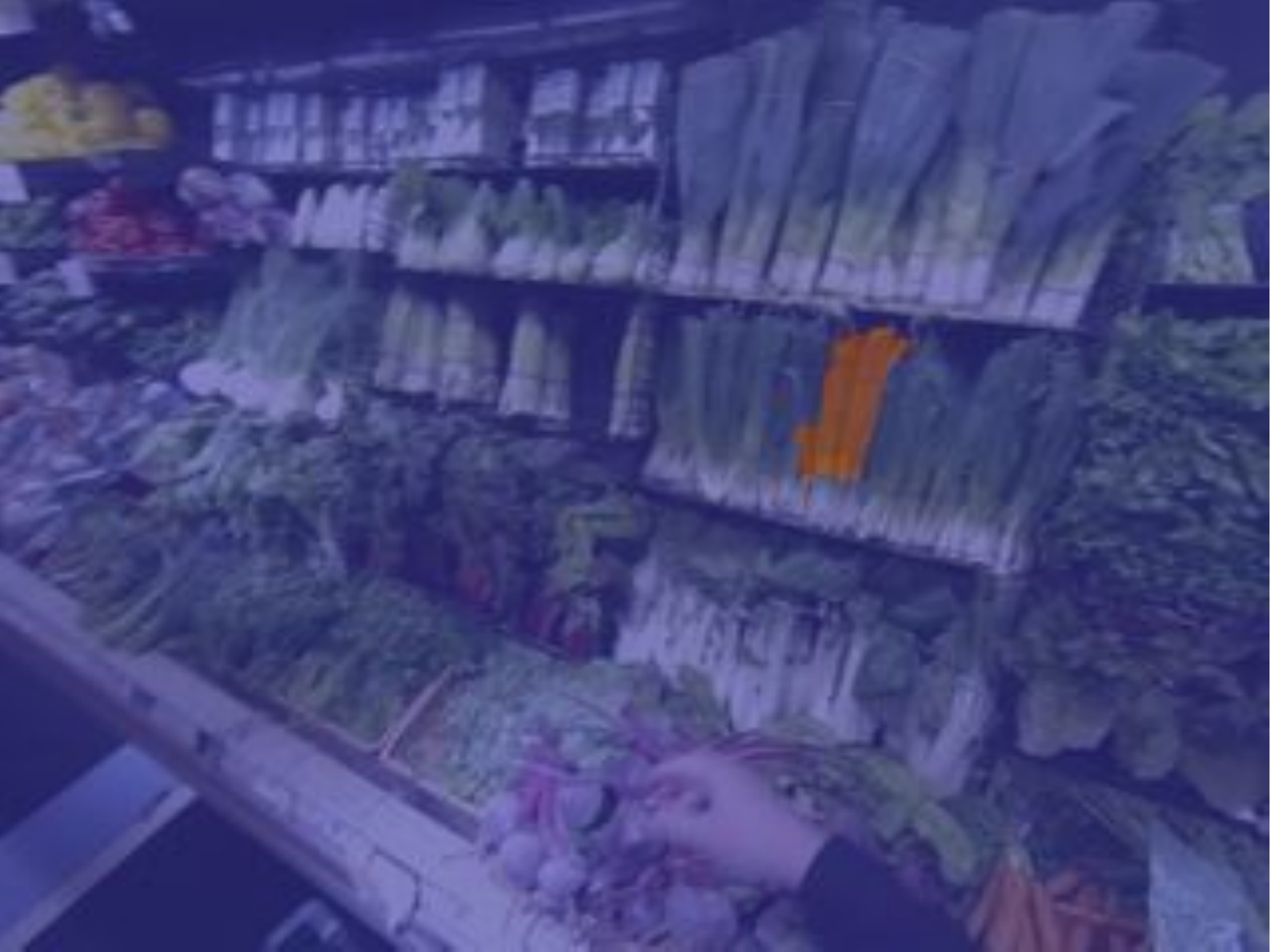}

\myfigurethreecol{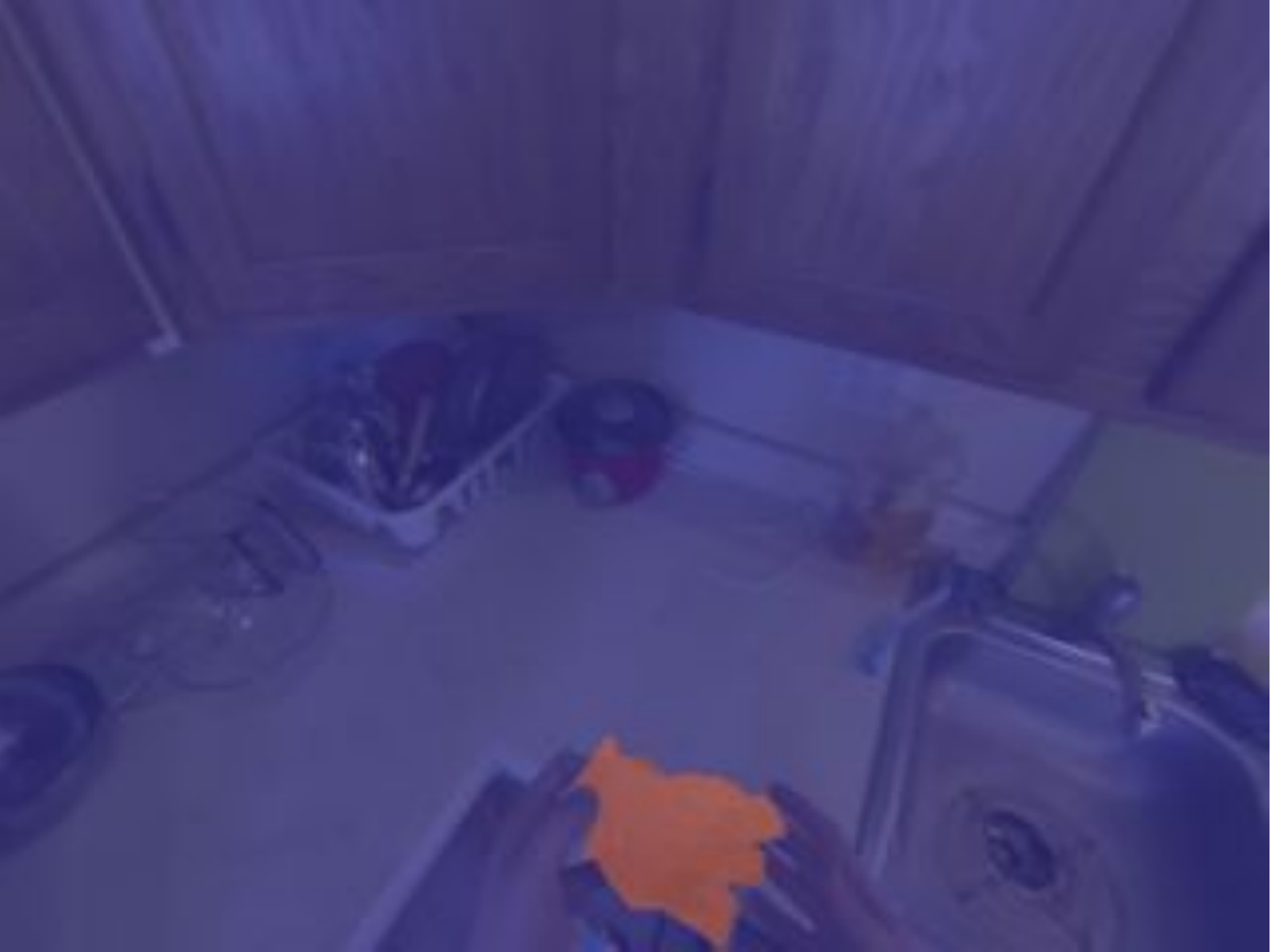}
\myfigurethreecol{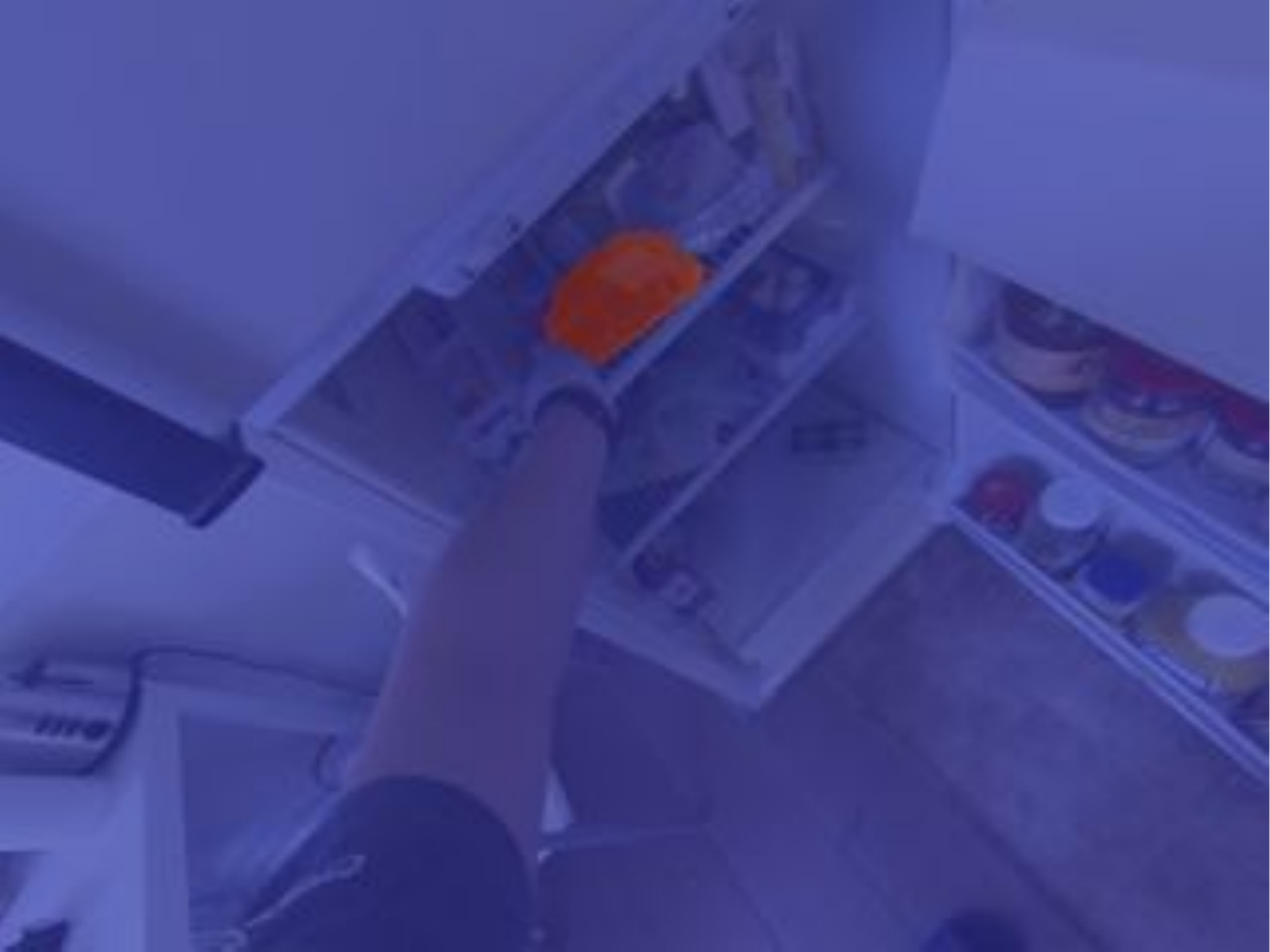}
\myfigurethreecol{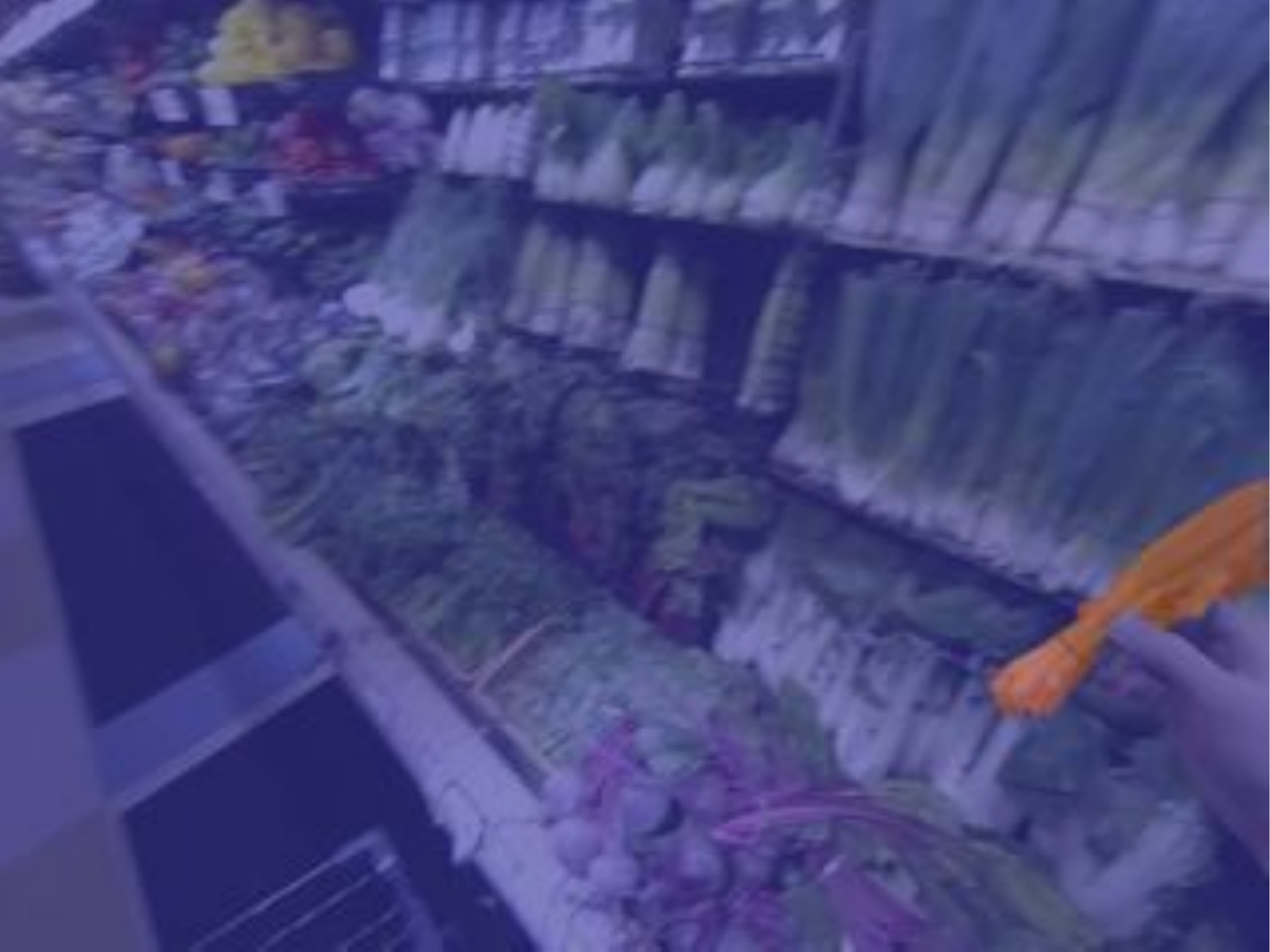}

\captionsetup{labelformat=default}
    \caption{ A few images from our Egocentric RGBD Saliency dataset illustrating the annotations for Future Saliency Prediction task (best viewed in color). In top row, we display frames corresponding to the present time overlaid with a non-salient object. The frames on the bottom illustrate the corresponding frames after $t$ seconds, where the same object is salient. Our goal here is to predict an object that will be salient after $t$ seconds.}
    \label{fig:data_fut}
\end{figure}

\textbf{Interaction Classification.} Most of the current computer vision systems classify objects by specific object class templates (cup, phone, etc). However, these templates are not very informative and cannot be used effectively beyond the tasks of object detection. Adding object's function, and the type of interaction related to that object would allow researchers to tackle a wider array of problems overlapping vision and psychology.


To predict an interaction type at a given frame, for each frame we select top $15$ highest ranked regions $R^{top}_{1 \hdots 15}$ according to their predicted saliency score. We then classify each of these regions either as sight or touch. Finally, we take the majority label from these $15$ classification predictions, and use it to classify an entire frame as sight or touch.




\captionsetup{labelformat=default}
\captionsetup[figure]{skip=10pt}

\setlength{\tabcolsep}{2pt}

     \begin{table*}
     \footnotesize
    \begin{center}
    \begin{tabular}{ | c | c | c | c | c | c | c | c | c | c | c | c | c | c | c | c | c ? c | c |}
    \hline
    \multirow{2}{*}{Method} & \multicolumn{2}{c|}{kitchen} & \multicolumn{2}{c|}{cooking} & \multicolumn{2}{c|}{eating} & \multicolumn{2}{c|}{dishwashing} & \multicolumn{2}{c|}{supermarket} & \multicolumn{2}{c|}{hotel 1} & \multicolumn{2}{c|}{hotel 2} & \multicolumn{2}{c ?}{foodmart} & \multicolumn{2}{c |}{\bf mean}\\ \cline{2-19}
    	& MF & AP & MF & AP & MF & AP & MF & AP & MF & AP & MF & AP & MF & AP & MF & AP & MF & AP\\ \hline
	  FTS~\cite{Achanta_frequency-tunedsalient} & 7.3 & 0.6 & 8.9 & 1.3 & 10.6 & 1.2  & 5.8 & 0.5 & 4.5 & 1.1 & 17.2 & 2.0 & 20.0 & 2.5 & 5.0 & 0.7 & 9.9 & 1.2\\	
	MCG~\cite{APBMM2014} & 10.4  & 4.7 & 13.8 & 7.0 & 21.1 & 12.7 & 7.1 & 2.9  & 12.5 & 5.7 & 23.6 & 12.2 & 31.2 & 14.9 &  11.1 & 5.1 & 16.4 & 8.1\\
	GBMR~\cite{yang2013saliency} & 8.0 & 3.0 & 15.6 & 6.8 & 14.7 & 7.0 & 6.8 & 3.0 & 4.3 & 1.3 & 32.7 & 18.3 & 46.0 & 30.2 & 12.9 & 5.7 & 17.6 & 9.4\\
	salObj~\cite{DBLP:journals/corr/LiHKRY14} & 7.2 & 2.7 & 19.9 & 7.4 & 21.3 & 10.0 & 15.4 & 5.1 & 5.8 & 2.2 & 24.1 & 9.2 & 49.3 & 28.3 & 9.0 & 3.4 & 19.0 & 8.5 \\
	GBVS~\cite{Harel07graph-basedvisual} & 7.2 & 3.0 & 21.3 & 11.4 & 20.0 & 10.6 & 16.1 & 8.8 & 4.3 & 1.4 &  23.1 & 13.8 & 50.9 & \bf 50.2 & 11.6 & 5.7 & 19.3 & 13.1 \\
	Objectness~\cite{objectness} & 11.5 & 5.6 & \bf 35.1 &  \bf 24.3 & 39.2 & 29.4 & 11.7 & 6.9 & 4.7 & 1.9 & 27.1 & 17.1 & 47.4 & 42.2 & 13.0 & 6.4 & 23.7 & 16.7 \\ \cline{1-19}
	\bf Ours (RGB) & 25.7 & 16.2 & 34.9 & 21.8 & 37.0 & 23.0 & 23.3 & 14.4 & \bf 28.9 & \bf 18.5 & 32.0 & 18.7 & \bf 56.0 & 39.6 & 30.3 & 21.8 & 33.5 & 21.7\\
	\bf Ours (RGB-D) & \bf 36.9 & \bf 26.6 & 30.6 & 18.2 & \bf 55.3 & \bf45.4 & \bf 26.8 & \bf 19.3 & 18.8 & 10.5 & \bf 37.9 & \bf 25.4 & 50.6 & 38.4 & \bf 40.2 & \bf 28.5 & \bf 37.1 & \bf 26.5\\ \hline
	 
    \end{tabular}
    \end{center}
    \caption{Our results for a 3D saliency detection task on our egocentric RGBD dataset. To evaluate the performance of each method we use the Max F-score (MF) and Average Precision (AP) metrics. Our method outperforms all prior saliency detection baselines by $13.4\%$ and $9.8\%$ according to MF and AP metrics respectively. Additionally, we note that using the depth information improves our method's accuracy by $3.6\%$ and $4.8\%$, which suggests that depth cues are important for egocentric saliency detection.}
    \label{po_table}
   \end{table*}

\section{Egocentric RGBD Saliency Dataset}
\label{data_sec}

We now present our Egocentric RGBD Saliency dataset. Our dataset records people's interactions with their environment from a first-person view in a variety of settings such as shopping, cooking, dining, etc. We use egocentric-stereo cameras to capture the depth of a scene as well. We note that in the context of our problem, the depth information is particularly useful because it provides an accurate distance from an object to a person. Since we hypothesize that a salient object from the 3D world maps to an egocentric RGBD frame with a predictable depth characteristic, we can use depth information as an informative cue for 3D saliency detection task.

Our dataset has annotations for three different tasks: saliency detection, future saliency prediction, and interaction classification. These annotations enables us to train our models in a supervised fashion and quantitatively evaluate our results. We now describe particular characteristics of our dataset in more detail.

\textbf{Data Collection.} We use two stereo GoPro Hero 3 (Black Edition) cameras with $100mm$ baseline to capture our dataset. All videos are recorded at $1280 \times 960$ with $100fps$. The stereo cameras are calibrated prior to the data collection and synchronized manually with a synchronization token at the beginning of each sequence. 

\textbf{Depth Computation.} We compute disparity between the stereo pair after stereo rectification. A cost space of stereo matching is generated for each scan line and match each pixel by exploiting dynamic programming in a coarse-to- fine manner.

\textbf{Sequences.} We record $8$ video sequences that capture people's interactions with object in a variety of different environments. These sequences include: cooking, supermarket, eating, hotel $1$, hotel $2$, dishwashing, foodmart, and kitchen sequences. 



\textbf{Saliency Annotations.} We use GrabCut software~\cite{Rother:2004:GIF:1015706.1015720} to annotate salient regions in our dataset. We generate $515$ annotated frames for kitchen, cooking, and eating sequences, $463$ and $646$ annotated frames for supermarket and foodmart sequences respectively, $410$ and $491$ annotated frames for hotel $1$ and hotel $2$ sequences respectively, and $674$ annotated frames for dishwashing sequence (for a total of $4229$ frames with per-pixel salient object annotations). In Fig.~\ref{fig:data_po}, we illustrate a few images from our dataset and the depth channels corresponding to these images. To illustrate ground truth labels, we overlay these images with saliency annotations.


%
%
%
%
%


Additionally, in Fig.~\ref{fig:data_stats}, we provide statistics that capture different properties of our dataset such as the location, depth, and size of annotated salient regions from all sequences. Each video sequence from our dataset is marked by a different color in this figure. We observe that these statistics suggest that different video sequences in our dataset exhibit different characteristics, and captures a variety of diverse interactions between people and objects. 




\textbf{Annotations for Future Saliency Prediction.} In addition, we also label our dataset to predict future saliency in egocentric RGBD image after $K$ frames. Specifically, we first find the frame pairs that are $K$ frames apart, such that the same object is present in both of the frames. We then check that this object is non-salient in the earlier frame and that it is salient in the later frame. Finally, we generate per-pixel annotations for these objects in both frames. We do this for the cases where the pair of frames are $2,4$, and $6$ seconds apart. We produce $48$ annotated frames for kitchen, $100$ for cooking, $42$ for eating, $164$ for supermarket, $48$ for hotel $1$, $29$ for hotel $2$, $31$ for foodmart, and $48$ frames dishwashing sequences. We present some examples of these annotations in Fig.~\ref{fig:data_fut}.

\textbf{Annotations for Interaction Classification.} To better understand the nature of people's interactions with their environment we also annotate each interaction either as {\em sight} or as {\em touch}.

\section{Experimental Results}
\label{exp_sec}


In this section, we present the results on our Egocentric RGBD Saliency dataset for three different tasks, which include 3D saliency detection, future saliency prediction and interaction classification. We show that using our EgoObject feature representation, we achieve solid quantitative and qualitative results for each of these tasks.

   \begin{figure}
\centering

\myfigurethreecol{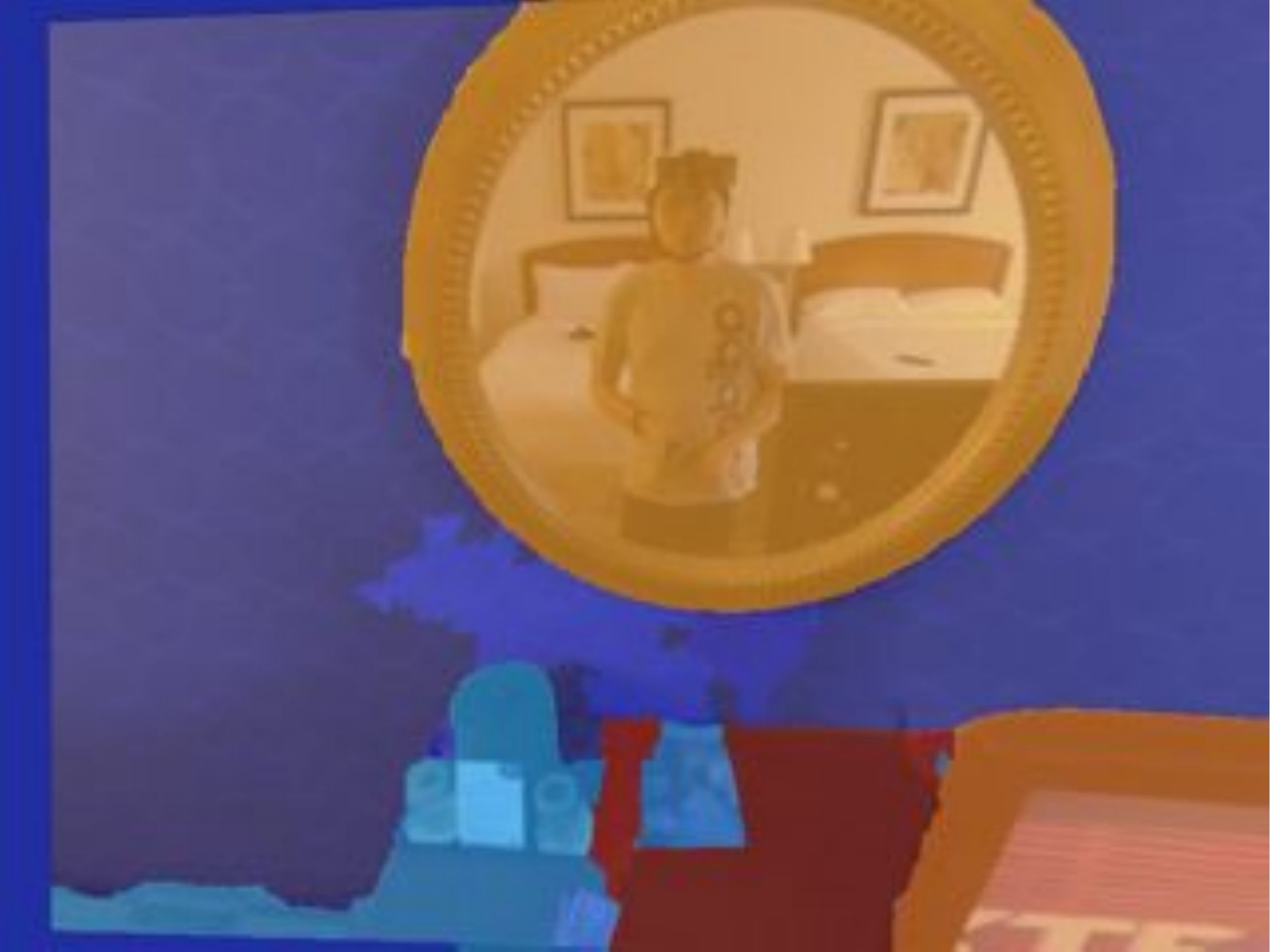}
\myfigurethreecol{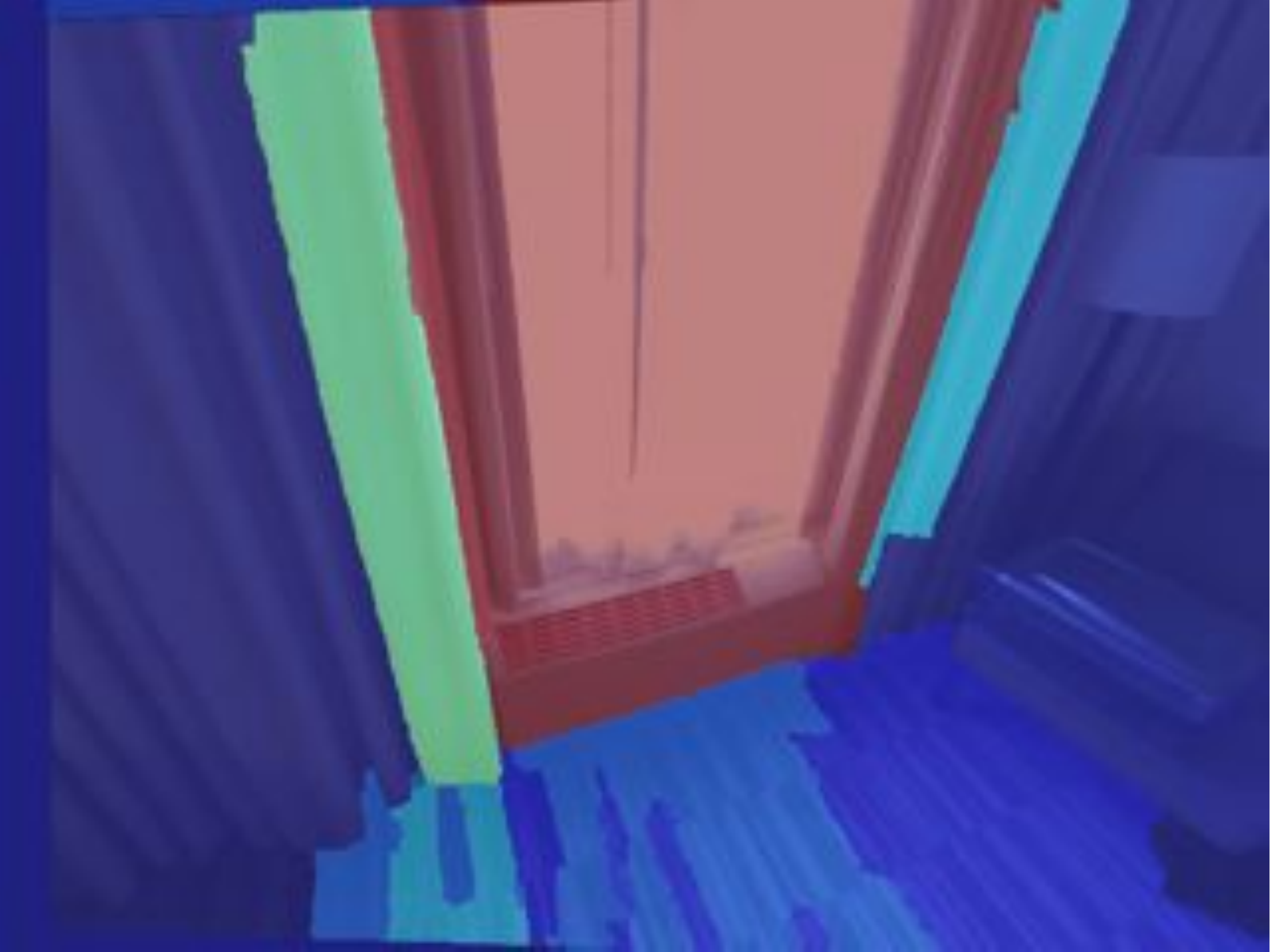}
\myfigurethreecol{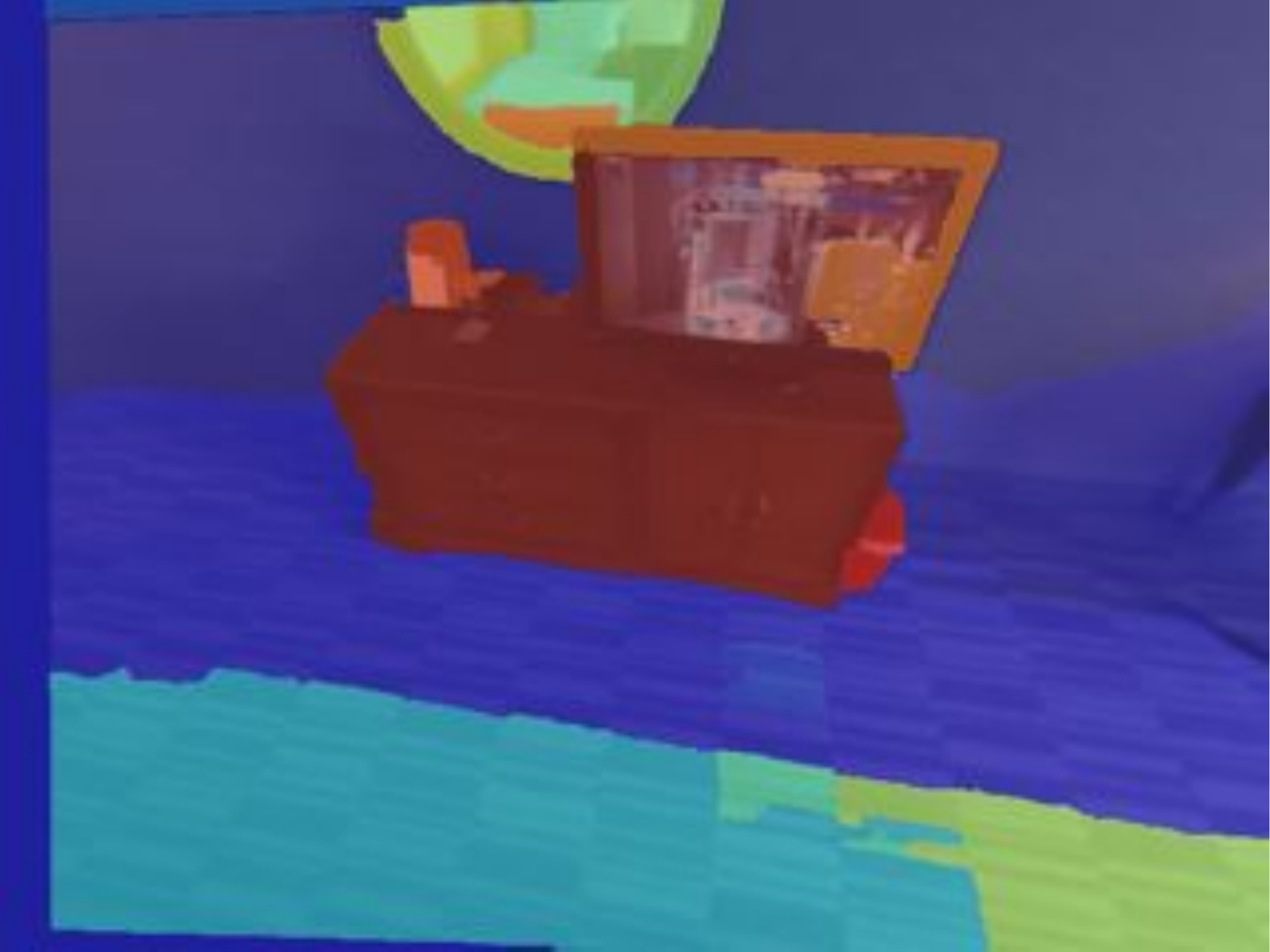}

\myfigurethreecol{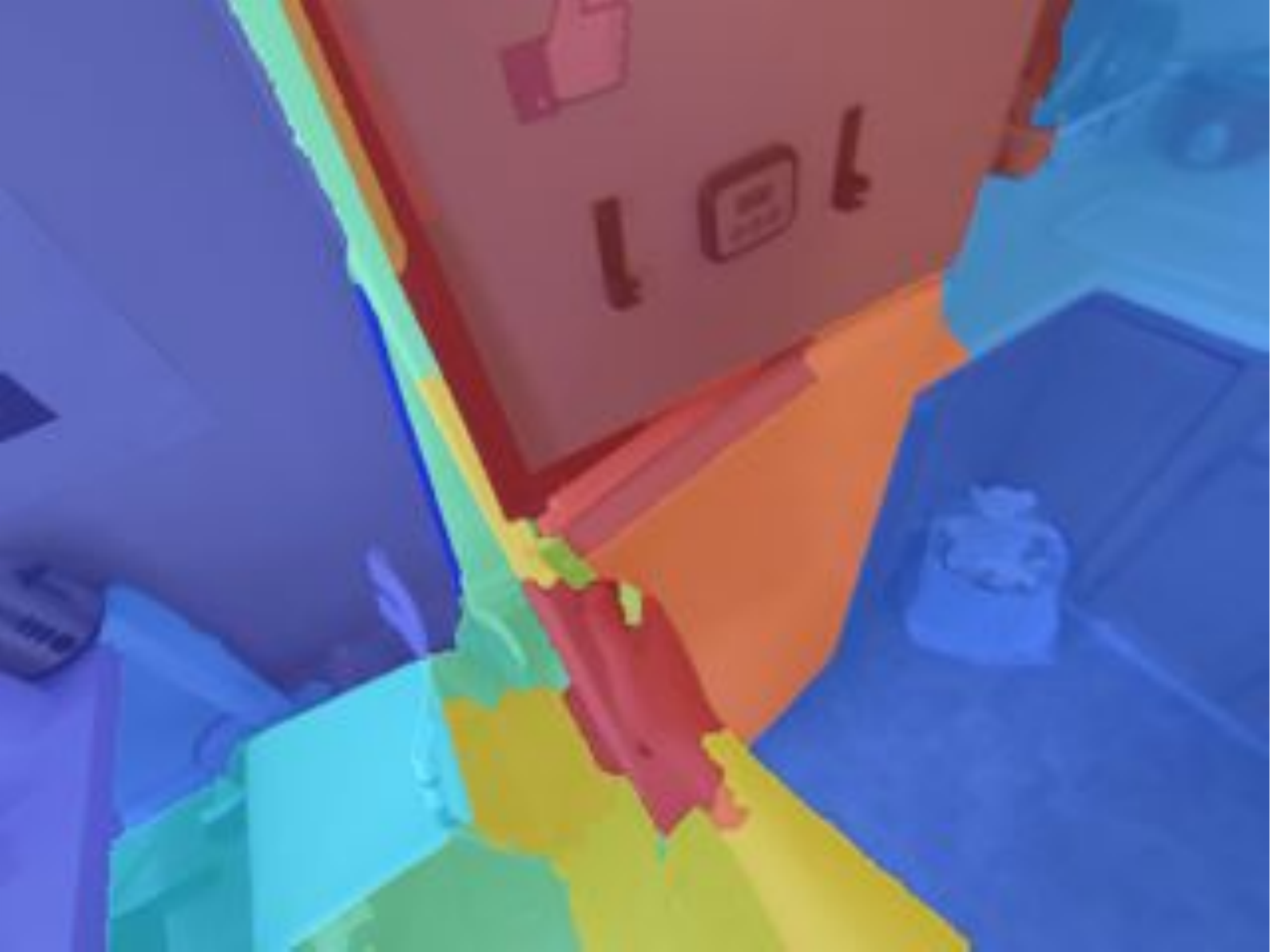}
\myfigurethreecol{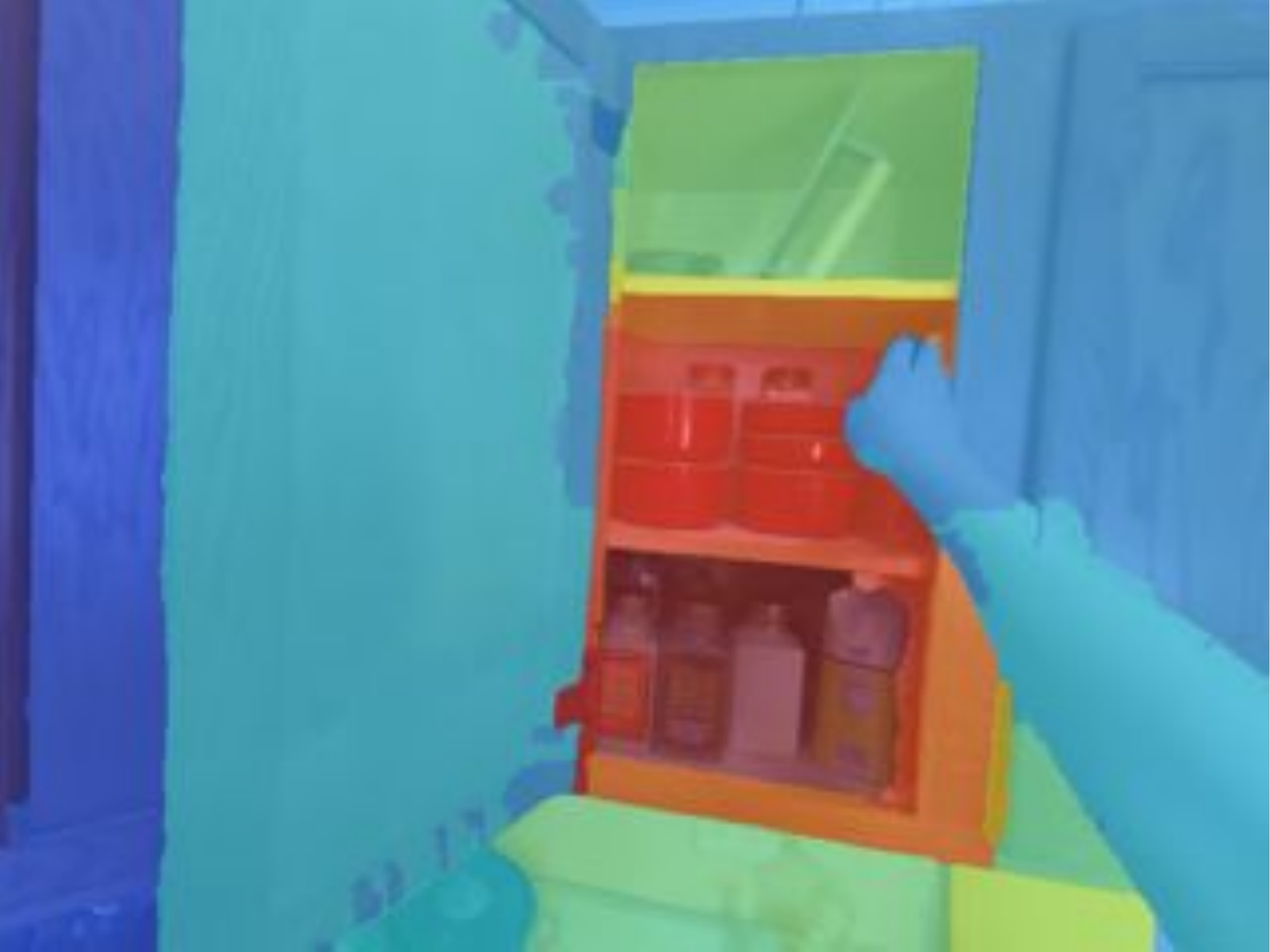}
\myfigurethreecol{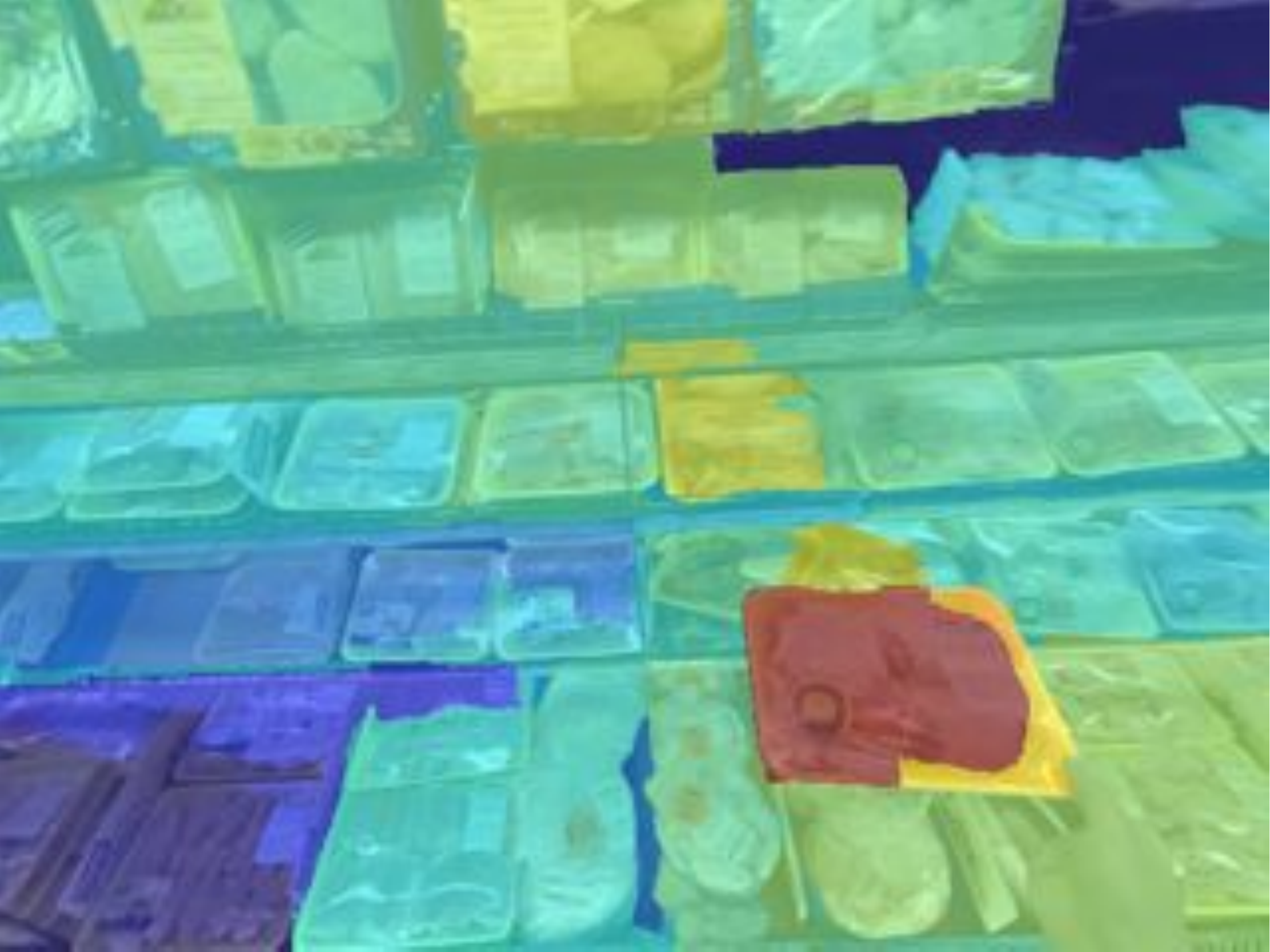}

\captionsetup{labelformat=default}
    \caption{An illustration of our qualitative results for a saliency detection and interaction classification tasks (best viewed in color). The first row depicts the interactions that are classified as \textbf{sight}, while the bottom row depicts \textbf{touch} interactions. The red color in the figure implies higher saliency, whereas the blue color denotes low saliency. Based on these results, we observe that using our approach we can accurately capture saliency in an egocentric RGBD frame and correctly distinguish between different types of people's interactions with objects. }
    \label{fig:po_preds}
\end{figure}

To evaluate our results, we use the following procedure for all three tasks. We first train random forest (RF) using the training data from $7$ sequences. We then use this trained RF to test it on the sequence that was not used in the training data. Such a setup ensures that our classifier is learning a meaningful pattern in the data, and thus, can generalize well on new data instances. We perform this procedure for each of the $8$ sequences separately and then use the resulting RF model to test on its corresponding sequence.

For the saliency detection and future saliency prediction tasks, our method predicts pixelwise saliency for each frame in the sequence.  To evaluate our results we use two different measures: a maximum F-Score (MF) along the Precision-Recall curve, and average precision (AP). For the task of interaction classification, we simply classify each interaction either as sight or as touch. Thus, to evaluate our performance we use the fraction of correctly classified predictions. We now present the results for each of these tasks in more detail.


\subsection{3D Saliency Detection}

Detecting 3D saliency in an Egocentric RGBD setting is a novel and relatively unexplored problem. Thus, we compare our method with the most successful saliency detection systems for 2D images. 



%
%
%
%

\begin{figure}
\centering

\myfigurethreecol{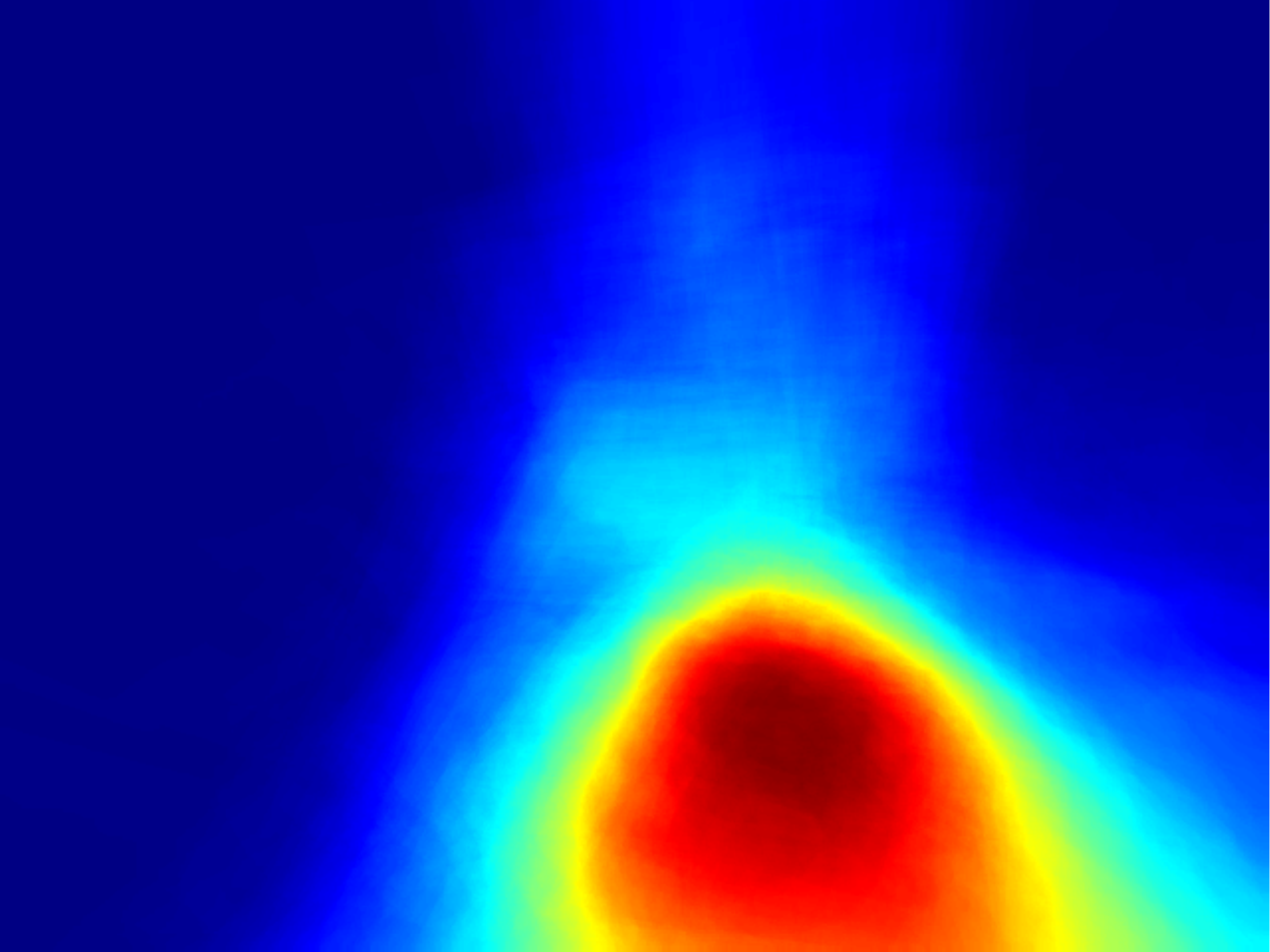}
\myfigurethreecol{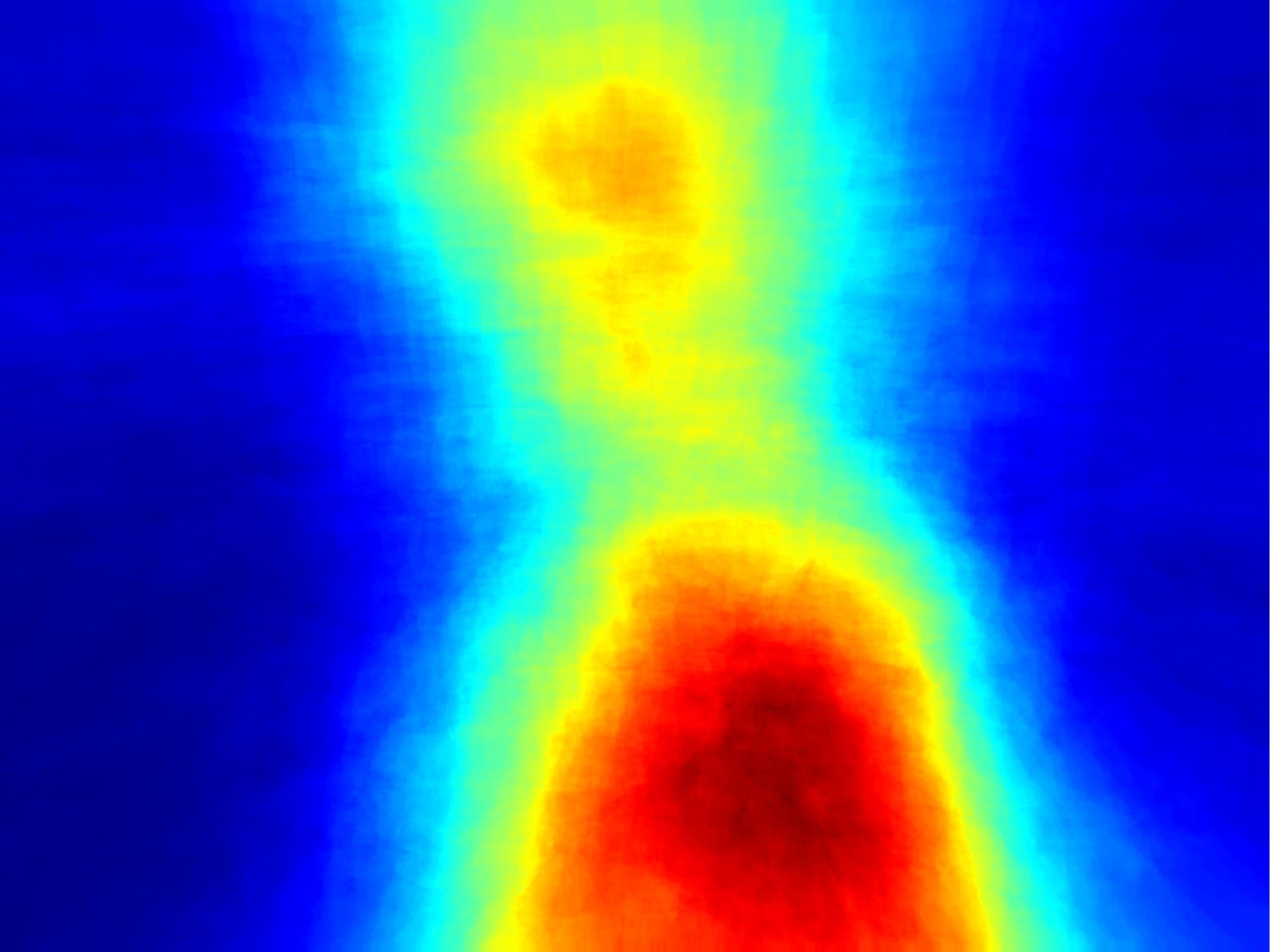}
\myfigurethreecol{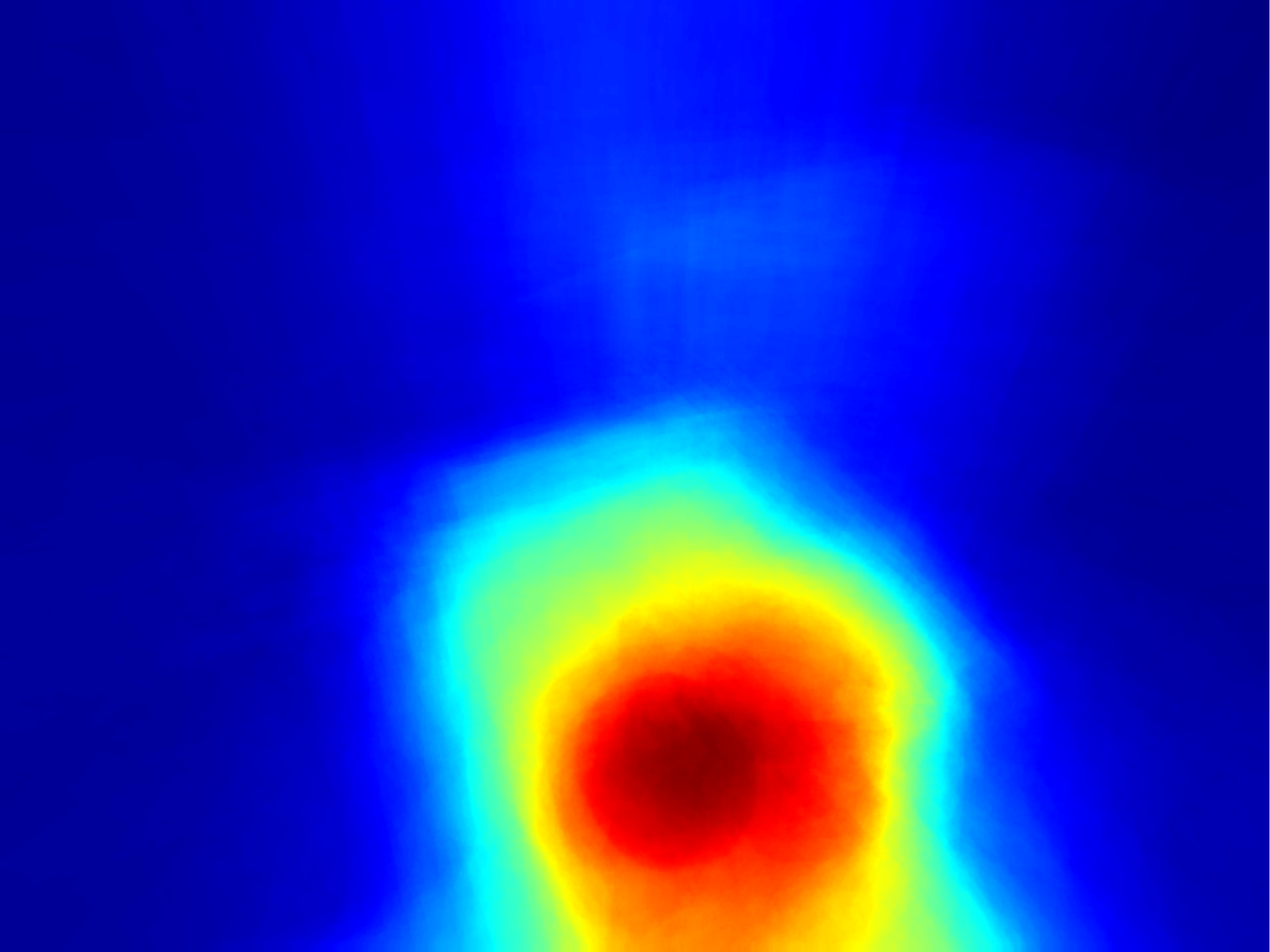}

\myfigurethreecol{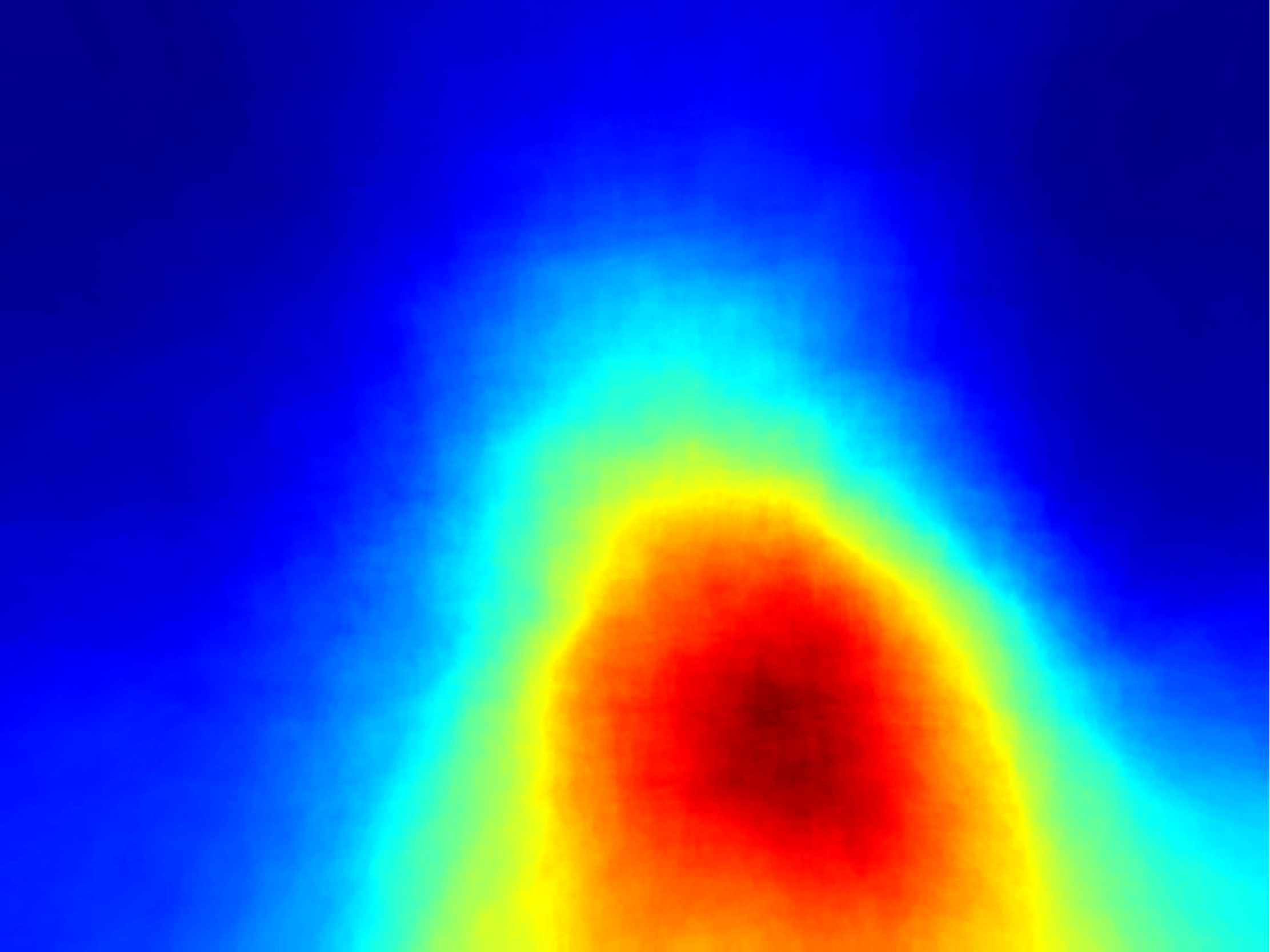}
\myfigurethreecol{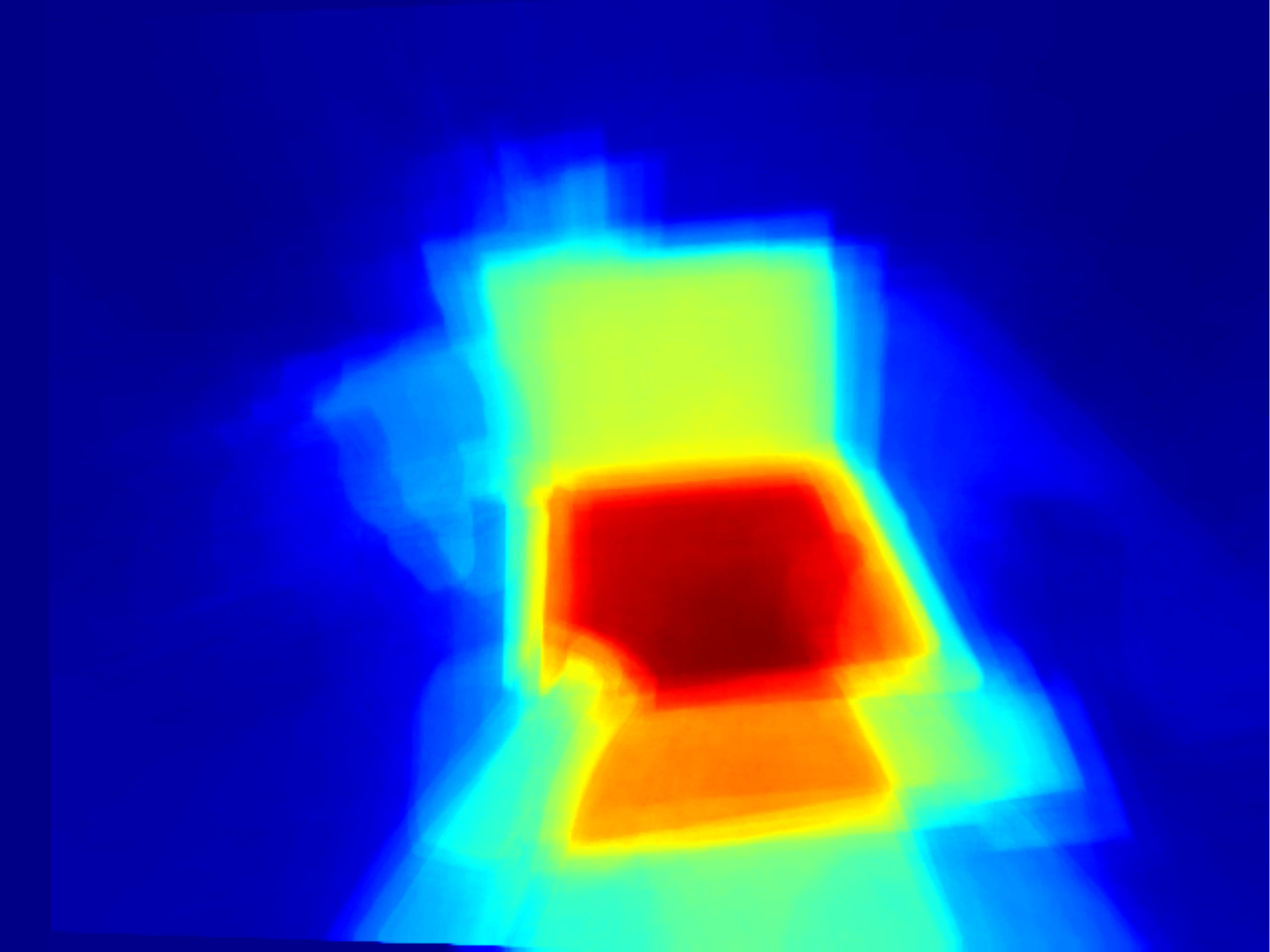}
\myfigurethreecol{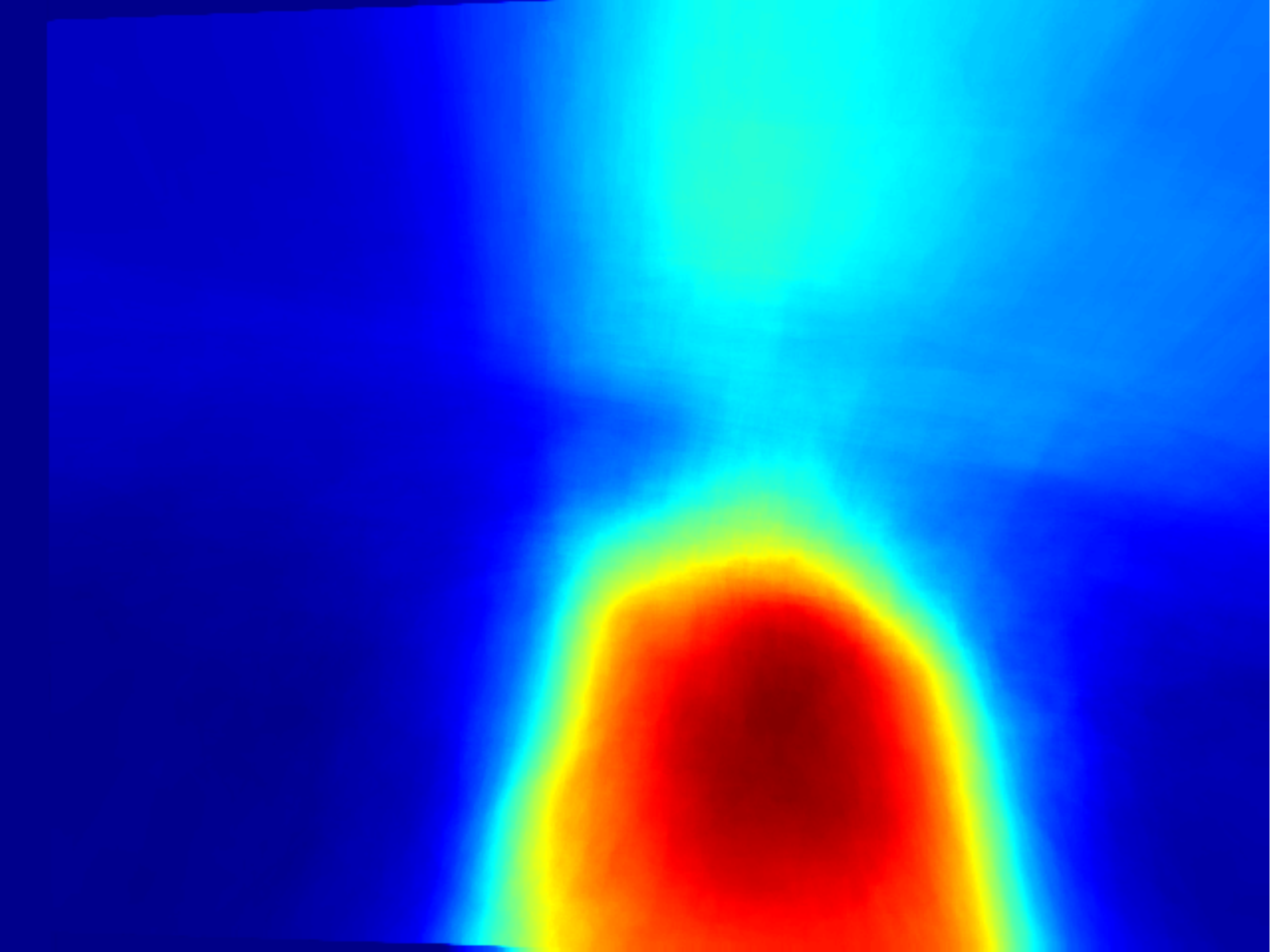}

\captionsetup{labelformat=default}
    \caption{ A figure that depicts the average saliency prediction heatmaps for the selected $6$ sequences. These visualizations demonstrate that in each of these sequences, our method captures an egocentric object prior that has a distinct shape, location, and size pattern.}
    \label{fig:avg_preds}
\end{figure}


In Table~\ref{po_table}, we present quantitative results for the saliency detection task on our dataset. We observe that our approach outperforms all the other methods by $13.4\%$ and $9.8\%$ in MF and AP evaluation metrics respectively. These results indicate that saliency detection methods designed for {\em non-egocentric} images do not generalize well to the {\em egocentric} images. This can be explained by the fact that in most {\em non-egocentric} saliency detection datasets, images are displayed at a pretty standard scale, with little occlusions, and also close to the center of an image. However, in the egocentric environment, salient objects are often occluded, they appear at a small scale and around many other objects, which makes this task more challenging.


Furthermore, we note that none of these baseline methods use depth information. Based on the results, in Table~\ref{po_table}, we observe that adding depth features to our framework provides accuracy gains of $3.6\%$ and $4.8\%$ according to MF and AP metrics respectively. Finally, we observe that the results of different methods vary quite a bit from sequence to sequence. This confirms that our Egocentric RGBD Saliency dataset captures various aspects of people's interactions with their environment, which makes it challenging to design a method that would perform equally well in each of these sequences. Based on the results, we see that our method achieves best results in $7$ and $6$ sequences (out of $8$) according to MF and AP evaluation metrics respectively, which suggests that exploiting egocentric object prior via shape, location, size, and depth features allows us to predict visual saliency robustly across all sequences.

Additionally, we present our qualitative results in Fig.~\ref{fig:po_preds}. Our saliency heatmaps in this figure suggest that we can accurately capture different types of salient interactions with objects. Furthermore, to provide a more interesting visualization of our learned egocentric object priors, we average our predicted saliency heatmaps for each of the $6$ selected sequences and visualize them in Fig.~\ref{fig:avg_preds}. We note that these averaged heatmaps have a certain shape, location, and size characteristics, which suggests the existence of an egocentric object prior in egocentric RGBD images.



\subsection{Feature Analysis}

\begin{figure}
\begin{center}
   \includegraphics[width=0.8\linewidth]{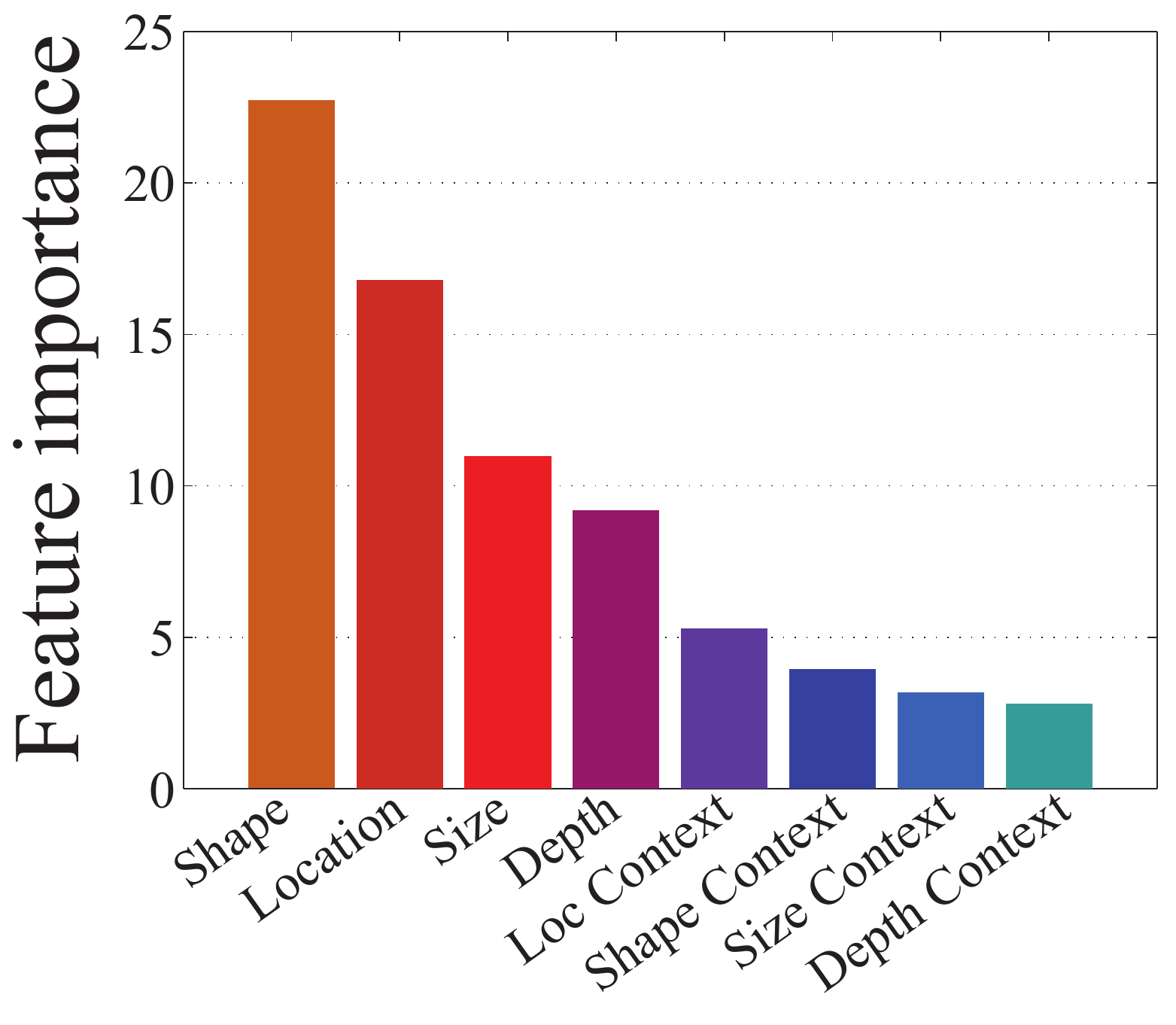}
\end{center}
\caption{A figure illustrating how various features contribute to the saliency detection task. Feature importance is evaluated by the mean squared error reduction when splitting the node by that feature in a random forest classifier. According to this evaluation, shape feature provide most informative cues, followed by the location, size and then depth features. As expected, the context features are least important.}
\label{fig:feats}
\end{figure}

In Fig.~\ref{fig:feats}, we also analyze, which features contribute the most for the saliency detection task. The feature importance is quantified by the mean squared error reduction when splitting the node by that feature in a random forest. In this case, we manually assign each of our $1089$ features to one of $8$ groups. These groups include shape, location, size, depth, shape context, location context, size context and depth context features (as shown in Fig.~\ref{fig:feats}). For each group, we average the importance value of all the features belonging to that group and present it in Figure~\ref{fig:feats}.

Based on this figure, we observe that shape features contribute the most for saliency detection. Additionally, since location features capture an approximate gaze of a person, they are deemed informative as well. Furthermore, we observe that size and depth features also provide informative cues for capturing the saliency in an egocentric RGBD image. As expected, the context feature are least important.

\captionsetup{labelformat=default}
\captionsetup[figure]{skip=10pt}

\subsection{Future Saliency Prediction Results}

In this section, we present our results for the task of future saliency prediction. We test our trained RF model under three scenarios: predicting a salient object that will be used after $2,4$, and $6$ seconds respectively. As one would expect, predicting the event further away from the present frame is more challenging, which is reflected by the results in Table~\ref{fut_table}. For this task, we aim to use our EgoObject representation to learn the cues captured by egocentric-stereo cameras that are indicative of person's future behavior. We compare our future saliency detector (FSD) to the saliency detector (SD) from the previous section and show that we can achieve superior results, which implies the existence and consistency of the cues that are indicative of person's future behavior. Such cues may include person's gaze direction (captured by an egocentric camera), or person's distance to an object (captured by the depth channel), which are both pretty indicative of what the person may do next.  

\begin{figure}
\centering

\myfigurethreecol{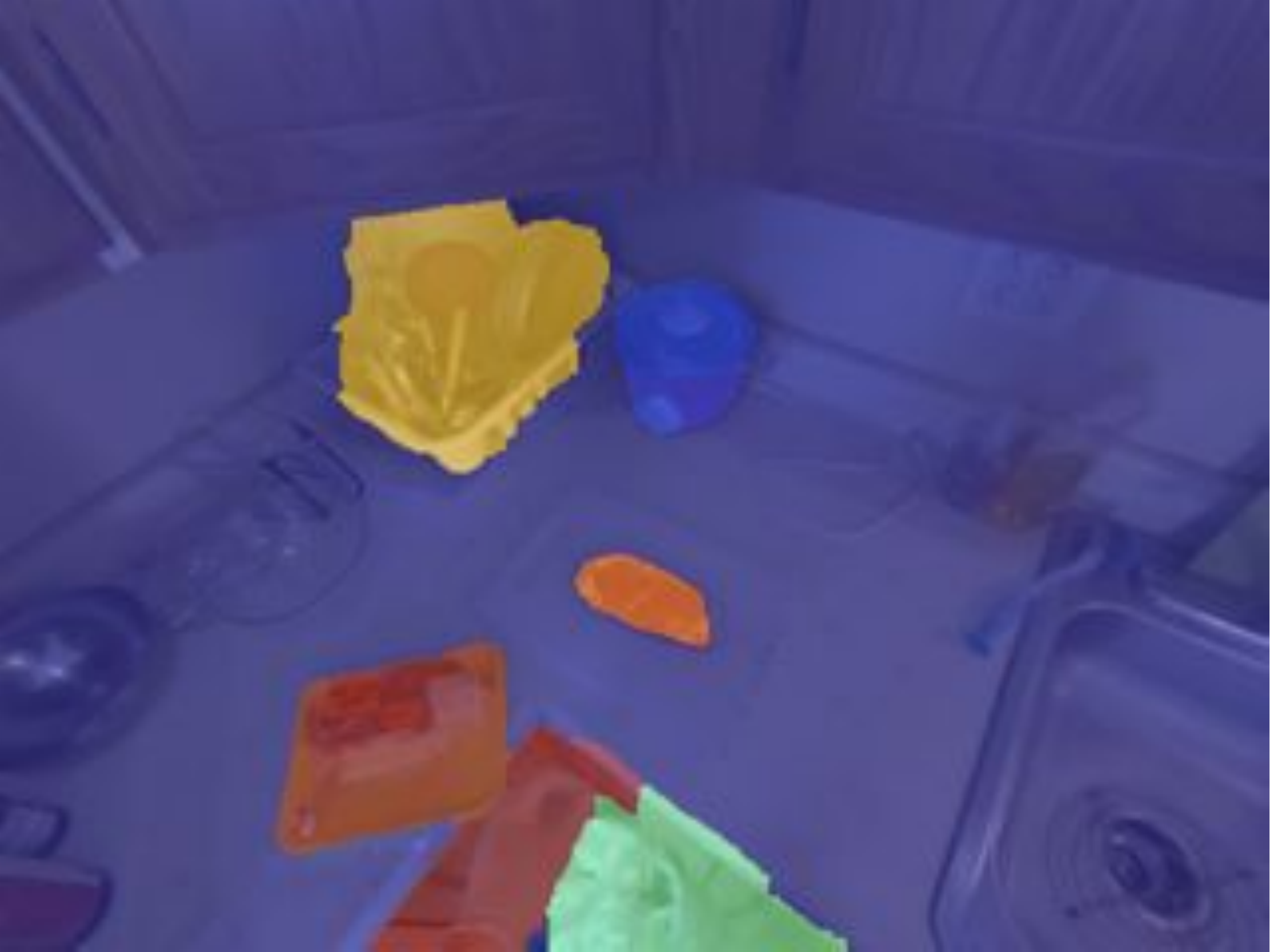}
\myfigurethreecol{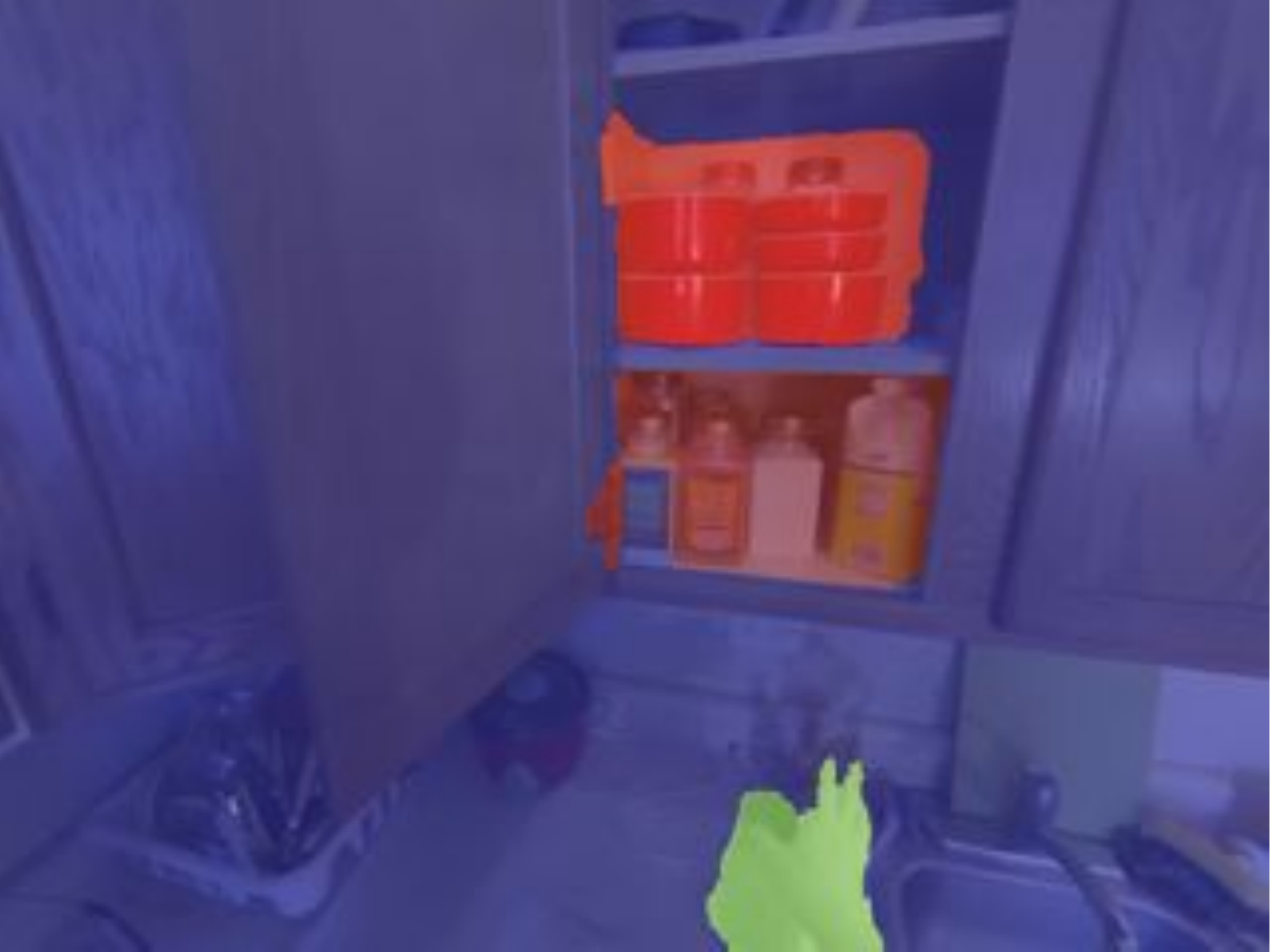}
\myfigurethreecol{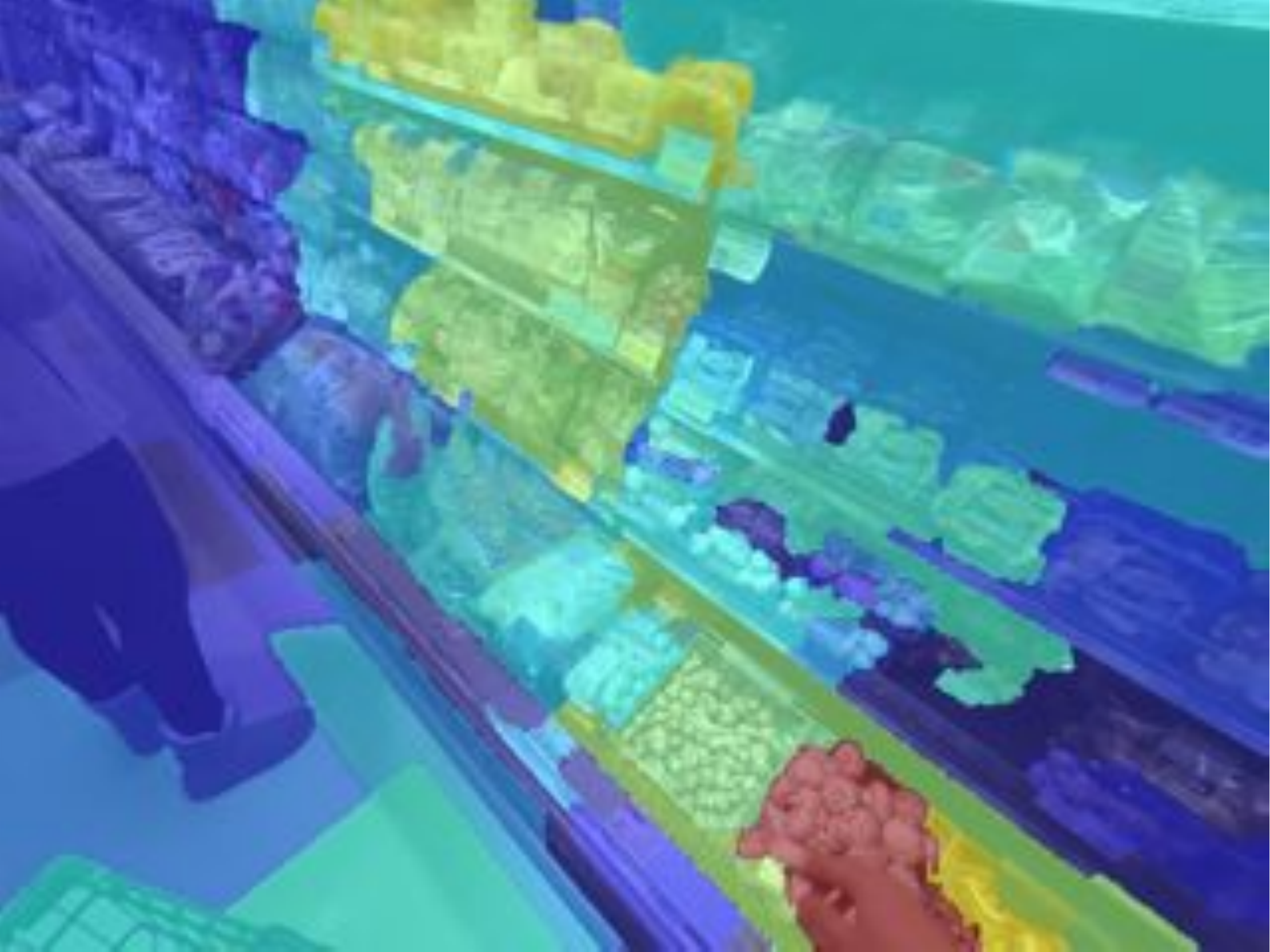}

\captionsetup{labelformat=default}
    \caption{ An illustration of our results for the future saliency detection task (best viewed in color). With a red color, we denote the regions that are likely to be salient in the future (e.g. a person will interact with them) while blue color indicates low future saliency values. We observe, that even in a difficult environment such as supermarket, our method produces meaningful future saliency predictions. }
    \label{fig:fut_preds}
\end{figure}

In Fig.~\ref{fig:fut_preds}, we visualize some of our future saliency predictions. Based on these results, we observe, that even in a difficult environment such as supermarket, our method can make meaningful predictions.


\captionsetup{labelformat=default}
  \begin{table}
  {\scriptsize
    \begin{center}
    \begin{tabular}{ | c | c | c | c | c |}
    \hline
    Metric & Method & 2 sec & 4 sec & 6 sec\\ \hline
     \multirow{2}{*}{MF}	
	& SD & 17.4 & 15.8 & 11.3\\ 
	& \bf FSD & \bf 26.1 & \bf 24.5 & \bf 22.1\\  \Xhline{4\arrayrulewidth}
     \multirow{2}{*}{AP}	
     	& SD & 7.3 & 7.4 & 4.7\\ 
	& \bf FSD & \bf 11.0 & \bf 10.8 & \bf 8.8\\ \hline
    \end{tabular}
    \end{center}
    \caption{Future saliency results according to Max F-score (MF) and Average Precision (AP) evaluation metrics. Given a frame at time $t$, our future saliency detector (FSD) predicts saliency for times $t+2,t+4$, and $t+6$ (denoted by seconds) . As our baseline we use a saliency detector (SD) from Section~\ref{tech_approach} of this paper. We show that in every case we outperform this baseline according to both metrics. This suggests that using our representation, we can consistently learn some of the egocentric cues such as gaze, or person's distance to an object that are indicative of people's future behavior.}
    \label{fut_table}}
   \end{table}

\subsection{Interaction Classification Results}

In this section, we report our results on the task of interaction classification. In this case, we only have two labels (sight and touch) and so we evaluate the performance as a fraction of correctly classified predictions. We compare our approach with a depth-based baseline, for which we learn an optimal depth threshold for each sequence, Then for a given input frame, if a predicted salient region is further than this threshold, our baseline classifies that interaction as {\em sight}, otherwise the baseline classifies it as {\em touch}. Due to lack of space, we do not present the full results. However, we note  that our approach outperforms depth-based baseline in $6$ out of $8$ categories and achieves $9.7\%$ higher accuracy on average in comparison to this baseline. We also illustrate some of the qualitative results in Fig.~\ref{fig:po_preds}. These results indicate that we can use our representation to successfully classify people's interactions with objects by sight or touch.


\section{Conclusion}


In this paper, we introduced a new psychologically inspired approach to a novel 3D saliency detection problem in egocentric RGBD images. We demonstrated that using our psychologically inspired EgoObject representation we can achieve good results for the three following tasks: 3D saliency detection, future saliency prediction, and interaction classification. These results suggest that an egocentric object prior exists and that using our representation, we can capture and exploit it for accurate 3D saliency detection on our egocentric RGBD Saliency dataset.

To conclude, we hope that our work in this paper, will contribute not only to the area of 3D saliency detection, but also to the broader studies concerned with human visual saliency perception.




\bibliographystyle{plain}
\footnotesize{
\bibliography{gb_bibliography}}

\end{document}